\pdfoutput=1
\documentclass{article}

\usepackage{iclr2027_conference,times}
\usepackage[utf8]{inputenc}
\usepackage[T1]{fontenc}
\usepackage{amsmath}
\usepackage{amssymb}
\usepackage{amsthm}
\usepackage{booktabs}
\usepackage{multirow}
\usepackage{array}
\usepackage{colortbl}
\usepackage{xcolor}
\usepackage{algorithm}
\usepackage{algpseudocode}
\usepackage{graphicx}
\usepackage{subcaption}
\usepackage{float}
\usepackage{microtype}
\usepackage{hyperref}
\hypersetup{hidelinks}
\usepackage{url}
\usepackage{xspace}
\usepackage{tikz}
\usetikzlibrary{arrows.meta,positioning,calc}

\newtheorem{proposition}{Proposition}

\newcommand{\method}{CDIS\xspace}
\newcommand{\gsm}{GSM8K\xspace}
\newcommand{\asdiv}{ASDiv\xspace}
\newcommand{\TopK}{\operatorname{TopK}}
\definecolor{oursgreen}{RGB}{228,240,252}
\newcommand{\oursrow}{\rowcolor{oursgreen}}
\newcolumntype{L}[1]{>{\raggedright\arraybackslash}p{#1}}

\iclrfinalcopy

\title{Refreshing Less, Selecting Better:\\
Reusing Stale Gradient Features for\\
Efficient Influence-Based Data Selection}

\author{Jianchang Su\\
University of Connecticut
\And
Yifan Zhang\\
University of Connecticut
\And
Wei Zhang\\
University of Connecticut}

\date{}

\begin{document}
\raggedbottom
\maketitle
\lhead{Preprint}

\begin{abstract}
Gradient-based data selection methods such as LESS score each candidate example by the alignment between its gradient and the gradient of a target validation set, and the scoring stage dominates their cost because per-example gradient features must be recomputed whenever the model checkpoint changes. Across three selection seeds, two model families, two candidate pools, and two target tasks, gradient features cached at a post-warmup checkpoint and paired with fresh validation gradients preserve the ranking 40 optimizer steps later with Spearman correlation between 0.952 and 0.991, while the top-10\% subset they induce misses 10 to 22\% of the examples that full recomputation selects. We therefore propose Cached Diverse Influence Selection (\method), which recomputes gradient features for the top-ranked fraction $p$ of candidates under the stale scores, fits an affine calibration on the recomputed examples, and selects the final subset under source and length quotas. A refresh fraction at or above the selection fraction recovers the exact top-$k$ subset whenever the calibrated stale scores have bounded error, and the measured budget curves follow this rule: at $p=0.3$ the recovered top-$k$ subset coincides with full recomputation in every setting, allocating the same budget per stratum recovers the stratified subset at 0.92 to 1.00, and the wall-clock of the gradient stage drops by a factor of 3.5 to 3.6. Iterating the cache over four checkpoints keeps top-$k$ overlap at 0.98 or higher at 1.9 gradient features per example against 4 for full recomputation, and the agreement between stale and recomputed scores on the refreshed examples provides a free check for unsafe reuse. Downstream, unconstrained top-$k$ selection on influence scores collapses to a single data source and scores 13 points below random selection on \gsm. \method scores 12 points above random selection with paired confidence intervals that exclude zero and trails full recomputation by 4.3 points, one training-run standard deviation, at 3.4 times lower selection cost.
\end{abstract}

\section{Introduction}
\label{sec:intro}

Post-training of large language models increasingly relies on data selection to reach a target capability within a fixed compute budget \citep{albalak2024survey,zhou2023lima,liu2024deita}. Gradient-based selection methods such as LESS \citep{xia2024less} represent each candidate example by a projected gradient feature at a model checkpoint, represent the target task by validation gradients at the same checkpoint, and select the candidates whose features align with the target \citep{koh2017influence,pruthi2020tracin,park2023trak}, which ties the selected data to the model and to the target distribution.

The cost of these methods is concentrated in the scoring stage. Over a post-training run, the gradient cost of a selection pipeline factorizes into three quantities: the cost of one per-example gradient feature, the number of candidates that receive a feature, and the number of times the features are recomputed. Prior work on efficiency has concentrated on the first factor: random projection reduces the dimension of each feature \citep{park2023trak,xia2024less}, proxy models reduce the cost of computing it \citep{chen2026iprox}, and learned influence models replace the gradient computation \citep{yu2024mates}. The third factor recurs in practice, because LESS computes gradient features at four checkpoints of a warmup run and averages the scores, and iterative pipelines re-select as the model changes \citep{mirzasoleiman2020craig,killamsetty2021gradmatch}. This paper asks how much of that recomputation is necessary.

Our starting point is a measurement of how influence scores change between post-training checkpoints. When train-gradient features cached at one checkpoint are paired with validation gradients from a later checkpoint, the resulting scores preserve the later checkpoint's ranking with Spearman correlation between 0.952 and 0.991 across three selection seeds, two model families, two candidate pools, and two target tasks, while the top-10\% subset they induce misses 10 to 22\% of the examples that full recomputation selects. This gap is where partial recomputation pays off: the stale ranking identifies where the subset can change, and the remaining candidates keep their cached features.

\begin{figure}[t]
\centering
\includegraphics[width=\textwidth]{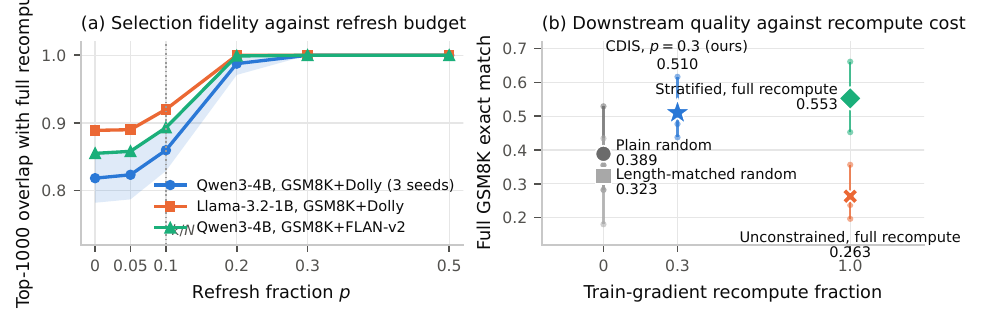}
\caption{Headline results. (a)~Overlap between the top-1,000 subset selected from partially refreshed scores and the subset from full recomputation against the refresh fraction $p$, for Qwen3-4B-Base on three \gsm+Dolly pools (mean and min--max band), Llama-3.2-1B with its own gradient features, and Qwen3-4B-Base on a \gsm+FLAN-v2 pool; the dotted line marks $k/N=0.1$. (b)~Full \gsm exact match against the fraction of train-gradient features recomputed per selection cycle, three-seed means with min--max whiskers.}
\label{fig:headline}
\end{figure}

We turn this observation into a selection procedure that we call Cached Diverse Influence Selection (\method). Given cached train-gradient features, fresh validation gradients, and a refresh fraction $p$, \method scores every candidate with the cached features, recomputes gradient features for the top $pN$ candidates under the stale scores, fits an affine map from stale to recomputed scores on those candidates, and assigns every remaining candidate its calibrated stale score. The budget has a lower bound with a reason: a refresh fraction at or above the selection fraction $k/N$ recovers the exact top-$k$ subset whenever the calibrated stale scores have bounded error (Proposition~\ref{prop:band}), and the measured budget curves rise steeply once $p$ passes $k/N$ and reach full fidelity at $p=0.3$ with 3.5 to 3.6 times lower wall-clock for the gradient stage (Figure~\ref{fig:headline}a). The same argument applied per stratum says how to allocate the budget for a stratified selection rule, and per-stratum allocation recovers the stratified subset at 0.92 to 1.00. The saving compounds when the cache is iterated, and over four checkpoints the iterated cache keeps top-$k$ overlap at 0.98 or higher at 1.9 gradient features per example against 4 for full recomputation. Finally, because the refreshed candidates carry both a stale and a recomputed score, their agreement comes free with the refresh, tracks the fidelity of the cached ranking, and separates unstable early-training intervals from stable ones.

A faithful cache is necessary for good selection, and the selection rule that consumes the scores matters as much. On a candidate pool that mixes 5,000 \gsm and 5,000 Dolly examples under a \gsm target, unconstrained top-$k$ selection on influence scores picks 1,000 \gsm examples and zero Dolly examples in every seed and scores 13 points below random selection on the full \gsm test set, because the selected subset collapses in source and length coverage. \method therefore applies the calibrated scores through a stratified selection rule with quotas over data sources and prompt-length buckets, and with this rule it scores 12.2 points above plain random selection and 24.7 points above unconstrained selection on full \gsm with paired confidence intervals that exclude zero, and 4.3 points, one training-run standard deviation, below stratified selection with full recomputation (Figure~\ref{fig:headline}b).

Our contributions are the following.
\begin{itemize}
\item We measure the temporal stability of influence-based train-gradient features across post-training checkpoints on three seeds, two model families, two candidate pools, two target tasks, and checkpoint distances from 40 to 240 optimizer steps.
\item We propose targeted partial refresh with affine calibration, give a sufficient condition on the refresh fraction for exact recovery of the top-$k$ subset, derive from it a rule for $p$ and a per-stratum allocation of the budget, and provide an online check for unsafe reuse that comes free with the refresh. At $p=0.3$ the method recovers the full-recomputation subset in every setting and outperforms random refresh and checkpoint skipping.
\item We show that cached influence scores require a diversity-preserving selection rule to be useful downstream, quantify the cost of caching against training-run variance, and show that the refresh composes with quota-based, normalization-based, and label-free selection rules.
\end{itemize}

\section{Related work}
\label{sec:related}

\paragraph{Influence estimation and gradient-based selection.}
Influence functions \citep{koh2017influence} and their scalable approximations \citep{grosse2023influence,kwon2024datainf,choe2025logra} estimate the effect of a training example on a model prediction; TracIn \citep{pruthi2020tracin} accumulates gradient inner products over checkpoints, TRAK \citep{park2023trak} and datamodels \citep{ilyas2022datamodels} attribute predictions at scale through random projection, and DsDm \citep{engstrom2024dsdm} applies datamodels to pretraining data selection. LESS \citep{xia2024less} adapts this line to instruction tuning with Adam-aware projected gradients stored in a datastore and computed at several warmup checkpoints; \method operates on that datastore and changes when its features are recomputed.

\paragraph{Reducing the cost of gradient-based selection.}
Existing work lowers the cost of a single scoring pass: random projection \citep{park2023trak,xia2024less} reduces the dimension of each gradient feature, IProX \citep{chen2026iprox} constructs proxy models whose gradients preserve influence at lower cost, MATES \citep{yu2024mates} trains a small data influence model that is retrained periodically as the pretraining model changes, and GREATS \citep{wang2024greats} moves selection into the training loop with ghost inner products. \method reduces the number of features that a checkpoint-aware pipeline recomputes, a separate factor of the total cost, and the two factors multiply: proxy or projected gradients can be cached and refreshed with the same procedure. The setting in which this matters most is repeated selection, as in gradient-matching coresets that re-solve selection every epoch \citep{mirzasoleiman2020craig,killamsetty2021gradmatch}. \citet{wang2025temporal} show that influence depends on the training stage; we measure how much post-training rankings drift between nearby checkpoints and exploit the concentration of that drift at the top of the ranking.

\paragraph{Diversity in instruction-tuning data selection.}
Quality-ranked or influence-ranked selection tends to produce unbalanced subsets, and several methods enforce balance or diversity: BIDS normalizes influence within task groups \citep{dai2025bids}, G-DIG and TAGCOS cluster gradient features \citep{pan2024gdig,zhang2025tagcos}, QDIT and D3 combine quality with diversity objectives \citep{bukharin2024qdit,zhang2025d3}, and DEITA and InsTag use complexity and tag diversity \citep{liu2024deita,lu2024instag}; \citet{xia2025random} report that many selection methods score below random selection at scale, which our unconstrained baseline reproduces; our contribution on this side is the interaction with caching, since a faithful cache inherits the collapse of full recomputation.

\section{Problem setup}
\label{sec:setup}

\paragraph{Gradient-based selection.}
Let $\mathcal{D}=\{x_i\}_{i=1}^{N}$ be a candidate pool, $\mathcal{V}$ a validation set for the target task, and $\theta_t$ a model checkpoint. A LESS-style selector represents each candidate by a projected, optimizer-aware gradient feature $g_i^t\in\mathbb{R}^d$ computed at $\theta_t$, reduces the validation gradients at $\theta_t$ to a target direction $v^t$, and scores each candidate by $s_i^t=\langle g_i^t, v^t\rangle$. LESS computes these scores at several checkpoints $t_1,\dots,t_T$ of a warmup run and averages them with learning-rate weights, $S_i=\sum_{j} w_j\, s_i^{t_j}$, before selecting the top-$k$ candidates \citep{xia2024less}. Computing $g_i^t$ for all $N$ candidates is the expensive step; $v^t$ comes from a small validation set and is cheap to refresh at every checkpoint.

\paragraph{Cost model.}
A pipeline that scores at $T$ checkpoints computes $TN$ train-gradient features at a cost of $c$ each; if the features from the first checkpoint are cached and a fraction $p$ of them is recomputed at each later checkpoint, the cost becomes $Nc+(T-1)pNc$ and the saving approaches a factor of $1/p$ as $T$ grows. Projection and proxy models reduce $c$, and our work reduces $p$.

\paragraph{Stale reuse and fidelity.}
Let $\ell<r$ be two checkpoints, with features $g_i^\ell$ cached at $\ell$ and validation gradients $v^r$ available at $r$. Full recomputation scores candidates by $s_i^r=\langle g_i^r,v^r\rangle$, and stale reuse scores them by
\begin{equation}
    \tilde{s}_i=\langle g_i^\ell, v^r\rangle ,
    \label{eq:stale}
\end{equation}
and we measure how well $\tilde{s}$ preserves $s^r$ with the Spearman rank correlation over the pool and with the overlap $|\mathcal{S}\cap\mathcal{S}_{\mathrm{full}}|/|\mathcal{S}_{\mathrm{full}}|$ between the subset $\mathcal{S}$ that a selection rule produces from $\tilde{s}$ and the subset $\mathcal{S}_{\mathrm{full}}$ that the same rule produces from $s^r$.

\section{Method}
\label{sec:method}

\method has two parts: a targeted partial refresh (Sections~\ref{sec:method_refresh} to~\ref{sec:method_check}) turns cached features and a refresh budget into a calibrated score vector, and a stratified selection rule (Section~\ref{sec:method_select}) turns the score vector into a subset. Figure~\ref{fig:system} contrasts the pipeline with prior balanced influence selection, and Algorithm~\ref{alg:cdis} in Appendix~\ref{app:algorithm} gives the full procedure.

\begin{figure}[t]
\centering
\begin{subfigure}[b]{\textwidth}
\centering
\includegraphics[width=0.8\textwidth,page=1]{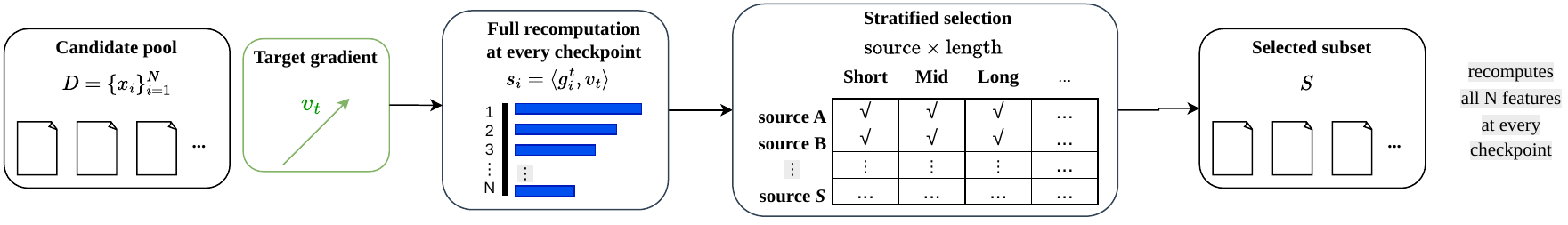}
\caption{Prior balanced influence selection scores every candidate at full cost at each checkpoint and then applies a stratified selection rule.}
\label{fig:system_prior}
\end{subfigure}
\vspace{0.4em}
\begin{subfigure}[b]{\textwidth}
\centering
\includegraphics[width=0.8\textwidth,page=2]{figures/overview_cropped.pdf}
\caption{\method scores the pool with cached features $C_{t-\Delta}$ and the fresh target gradient $v_t$, recomputes exact scores for the top-$pN$ refresh set $\mathcal{R}_p$, calibrates the remaining stale scores, writes the recomputed features back to the cache, and applies the same stratified selection rule to the updated scores $\hat{s}$.}
\label{fig:system_cdis}
\end{subfigure}
\caption{From prior balanced influence selection to \method. The selection rule on the right of each panel is shared; the refresh layer in (b) reduces the number of train-gradient features recomputed per checkpoint, and the stale and recomputed scores on $\mathcal{R}_p$ provide the online check of Section~\ref{sec:method_check}.}
\label{fig:system}
\end{figure}

\subsection{Targeted partial refresh}
\label{sec:method_refresh}

Given cached features $G^\ell=\{g_i^\ell\}_{i=1}^N$, the current validation direction $v^r$, and a refresh fraction $p\in(0,1]$, the refresh proceeds in four steps. First, it computes the stale score $\tilde{s}_i=\langle g_i^\ell, v^r\rangle$ for every candidate. Second, it forms the refresh set $\mathcal{R}_p=\TopK_{\lceil pN\rceil}([N];\tilde{s})$ from the $pN$ candidates with the highest stale scores. Third, it recomputes $g_i^r$ and the exact score $s_i^r=\langle g_i^r,v^r\rangle$ for every $i\in\mathcal{R}_p$ and writes $g_i^r$ back to the cache. Fourth, it fits an affine calibration
\begin{equation}
    (a,b)=\arg\min_{a,b}\sum_{i\in\mathcal{R}_p}\bigl(a\tilde{s}_i+b-s_i^r\bigr)^2
    \label{eq:calib}
\end{equation}
on the refreshed candidates and returns the hybrid scores $\hat{s}_i=s_i^r$ for $i\in\mathcal{R}_p$ and $\hat{s}_i=a\tilde{s}_i+b$ otherwise. The refresh set is chosen by stale rank because the candidates that can enter or leave the selected subset after recomputation are those whose stale scores lie near the selection threshold, which for a top-$k$ rule with $k\le pN$ sit at the top of the stale ranking. The calibration places recomputed and stale scores on a common scale, which matters whenever a selection rule compares candidates inside and outside $\mathcal{R}_p$, and the next checkpoint repeats the procedure on the updated cache.

\subsection{A sufficient refresh budget}
\label{sec:method_theory}

\begin{proposition}[Sufficient refresh set]
\label{prop:band}
Let $s^r$ be the recomputed scores, assumed distinct, let $\mathcal{T}_k$ be the top-$k$ subset under $s^r$, and let $\tau=\min_{i\in\mathcal{T}_k}s_i^r$. Let $c(\tilde{s}_i)=a\tilde{s}_i+b$ with $a>0$ be the calibrated stale scores and let $\varepsilon=\max_i|s_i^r-c(\tilde{s}_i)|$ be their maximum error. If the refresh set satisfies $\mathcal{R}\supseteq\{i: c(\tilde{s}_i)\ge\tau-\varepsilon\}$, then the top-$k$ subset under the hybrid scores $\hat{s}$ equals $\mathcal{T}_k$.
\end{proposition}

The proof is in Appendix~\ref{app:proof}. Because $\mathcal{T}_k$ is contained in the set $\{i: c(\tilde{s}_i)\ge\tau-\varepsilon\}$, that set has at least $k$ elements, so $p\ge k/N$ is necessary for the condition, and the number of additional candidates it requires grows with the error $\varepsilon$ and with the density of stale scores just below the threshold. This gives a rule for $p$ before any downstream training: the selection fraction plus a margin for the drift, widened for long or early intervals. The error $\varepsilon$ depends on the whole pool, and the refresh set exposes the residuals on its members, which the check of Section~\ref{sec:method_check} reads. For a stratified rule the same argument applies within each stratum, with the stratum's own threshold and quota, so exact recovery requires the refresh set to contain the union of the per-stratum margin sets. A single refresh set chosen by global stale rank covers the strata whose scores are high under the target and can leave strata with low scores, such as general instruction data under a mathematical target, to their calibrated stale scores. We therefore consider two allocations of the same budget $pN$: global allocation refreshes the top-$pN$ candidates of the pool, and per-stratum allocation refreshes the top-$p$ fraction of every stratum, which covers each stratum's margin whenever $p$ exceeds the stratum's own selection fraction. Section~\ref{sec:results_fidelity} confirms the prediction: global allocation recovers the top-$k$ subset exactly and the stratified subset at 77 to 94\%, with the missing examples in the lower-scoring source, and per-stratum allocation recovers the stratified subset at 92 to 100\%.

\subsection{An online check for unsafe reuse}
\label{sec:method_check}

The candidates in $\mathcal{R}_p$ carry both a stale and a recomputed score, so their agreement is observable at refresh time as a by-product of the refresh. We use the Spearman correlation between $\tilde{s}$ and $s^r$ on $\mathcal{R}_p$ as a check: low agreement indicates that the cached checkpoint lies in an unstable region of training or too far from the current checkpoint, and a practitioner then increases $p$ or recomputes all features. Section~\ref{sec:results_distance} validates this statistic on every measured interval.

\subsection{Stratified selection}
\label{sec:method_select}

The final subset is selected from $\hat{s}$ under quotas. Source quotas split $k$ across data sources. Within each source, candidates are split into four equal-count buckets by prompt length, and the source quota is divided evenly across buckets with largest-remainder rounding. Within each stratum the highest-scoring candidates fill the quota, and underfilled strata are topped up from the same source and then from the whole pool by score. The main experiments use two sources with equal quotas, the uniform allocation over sources (Section~\ref{sec:results_rules}). When source labels are unavailable, we cluster the cached gradient features with $k$-means and allocate quotas in proportion to cluster size.

\section{Experimental setup}
\label{sec:exp}

\paragraph{Models, warmup, and pools.}
The main model is Qwen3-4B-Base \citep{yang2025qwen3}; Llama-3.2-1B \citep{grattafiori2024llama3,meta2024llama32} serves as a second model family with its own gradient features, warmup, and selections. Following LESS, each model is warmed up with LoRA \citep{hu2022lora} on a random 7\% of the Dolly instruction data \citep{databricks2023dolly} for one epoch of 132 optimizer steps, with checkpoints every 40 steps. Cached features come from checkpoint 80 and the selection checkpoint is 120; the early intervals 40$\to$80 and 40$\to$120 serve as unstable comparisons, and a three-epoch warmup with the same data and spacing provides checkpoints 160, 240, and 320 for the distance and repeated-selection studies. The main pool contains 5,000 \gsm training examples \citep{cobbe2021gsm8k} and 5,000 Dolly examples, sampled per selection seed (seeds 0, 1, and 2); a second pool replaces Dolly by 5,000 FLAN-v2 examples \citep{longpre2023flan}. The main target is \gsm, with the validation direction from the 100 \gsm validation examples of the LESS pipeline; the MMLU target \citep{hendrycks2021mmlu} uses the 285 MMLU validation examples of LESS with its per-subtask reduction. Selection takes the top 10\%, that is, $k=1{,}000$ examples.

\paragraph{Features, training, and evaluation.}
Train and validation features are Adam-aware projected gradients from the LESS implementation with $d=8{,}192$, collected at checkpoints 40, 80, and 120. Each selected subset trains a LoRA adapter for five epochs with the warmup hyperparameters (Appendix~\ref{app:impl}), and the same adapters are evaluated with the LESS 8-shot evaluator on the full \gsm test set (1,319 items) and on the 2,232 numeric-answer items of \asdiv \citep{miao2020asdiv}, with exact match after normalization as the metric. Absolute accuracies under this protocol sit below few-shot results reported for the base model, since adapters train on chat-formatted examples and are scored by exact match within 256 generated tokens, and every method shares it. Selection-rule ablations use a 200-item \gsm subset (GSM200). All methods answer the same test items, so we report paired, seed-stratified bootstrap confidence intervals over per-item correctness differences (10,000 resamples), and we retrain four fixed selections with three training seeds each to measure the within-selection standard deviation.

\paragraph{Baselines and comparisons.}
Downstream baselines are unconstrained top-$k$ with full recomputation, plain random selection, length-matched random selection (matching the prompt-length profile of the unconstrained subset), and stratified selection with full recomputation. Fidelity comparisons at equal or larger savings are random refresh (a random 30\% of the pool) and checkpoint skipping (features from a checkpoint twice as far back, with the refresh skipped). BIDS-style normalization \citep{dai2025bids} and a subset-transfer diagnostic on Mistral-7B-v0.3 \citep{jiang2023mistral,mistral2024v03} complete the comparisons (Appendix~\ref{app:mistral}), and wall-clock is measured on one NVIDIA A6000 (48\,GB).

\section{Results}
\label{sec:results}

\subsection{Stale reuse and targeted refresh}
\label{sec:results_fidelity}

\begin{table}[t]
\centering
\scriptsize
\setlength{\tabcolsep}{3.4pt}
\begin{tabular}{lcccccccccc}
\toprule
& & \multicolumn{3}{c}{Top-1,000 overlap} & \multicolumn{3}{c}{Stratified, global allocation} & \multicolumn{2}{c}{Stratified, per-stratum} \\
\cmidrule(lr){3-5}\cmidrule(lr){6-8}\cmidrule(lr){9-10}
Setting & Spearman & stale & $p{=}0.2$ & $p{=}0.3$ & stale & $p{=}0.2$ & $p{=}0.3$ & $p{=}0.2$ & $p{=}0.3$ \\
\midrule
Qwen3-4B, seed 0 & 0.987 & 0.856 & 0.999 & 1.000 & 0.824 & 0.920 & 0.927 & 0.986 & 0.999 \\
Qwen3-4B, seed 1 & 0.976 & 0.782 & 0.971 & 1.000 & 0.720 & 0.837 & 0.857 & 0.955 & 0.996 \\
Qwen3-4B, seed 2 & 0.988 & 0.817 & 0.993 & 1.000 & 0.830 & 0.935 & 0.940 & 0.986 & 0.998 \\
Llama-3.2-1B, seed 0 & 0.991 & 0.889 & 1.000 & 1.000 & 0.829 & 0.885 & 0.904 & 0.978 & 0.992 \\
Qwen3-4B, \gsm+FLAN-v2 pool & 0.989 & 0.855 & 0.999 & 1.000 & 0.833 & 0.929 & 0.936 & 0.989 & 1.000 \\
Qwen3-4B, seed 0, MMLU target & 0.969 & 0.872 & 0.996 & 1.000 & 0.733 & 0.819 & 0.819 & 0.934 & 0.977 \\
Qwen3-4B, seed 1, MMLU target & 0.952 & 0.896 & 0.996 & 1.000 & 0.698 & 0.765 & 0.765 & 0.864 & 0.919 \\
\bottomrule
\end{tabular}
\caption{Selection fidelity of stale reuse and targeted refresh (checkpoint 80 to 120; \gsm target on \gsm+Dolly pools unless noted). Overlap is on the top-10\% subset under unconstrained top-$k$ and under the stratified rule; global allocation refreshes the top-$pN$ candidates of the pool and per-stratum allocation the top-$p$ fraction of every stratum, at the same budget.}
\label{tab:fidelity}
\end{table}

\begin{figure}[t]
\centering
\includegraphics[width=\textwidth]{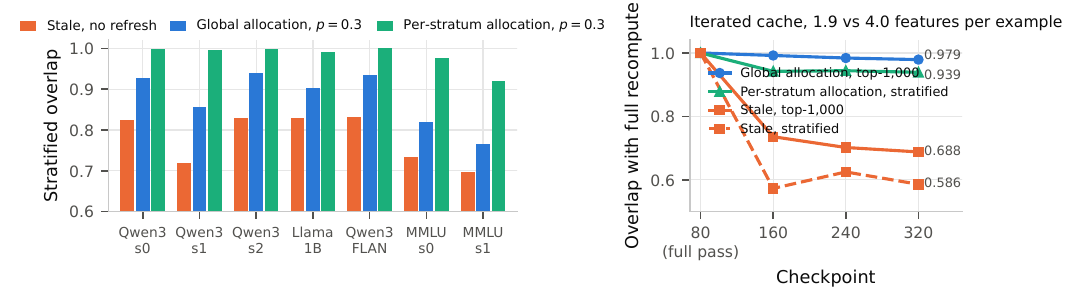}
\caption{(a)~Stratified overlap with full recomputation for stale reuse, global allocation, and per-stratum allocation at $p=0.3$ on the seven settings of Table~\ref{tab:fidelity}. (b)~Repeated selection with an iterated cache on the three-epoch run (Qwen3-4B seed 0): a full pass at checkpoint 80, then $0.3N$ features refreshed at each later checkpoint, at 1.9 features per example against 4.0 (Appendix~\ref{app:chain}).}
\label{fig:allocation_chain}
\end{figure}

Table~\ref{tab:fidelity} reports how much of the checkpoint-120 selection survives with checkpoint-80 features and fresh validation gradients. The global ranking survives almost intact, with Spearman correlations between 0.952 and 0.991 across all seven settings, and the selected subset shifts: stale scores recover 0.78 to 0.90 of the top-1,000 subset and 0.70 to 0.83 of the stratified subset. This is the situation that Proposition~\ref{prop:band} anticipates, because the drift is small relative to the score range and still large enough near the selection threshold to move up to 30\% of the selected examples.

\paragraph{Targeted refresh.}
Targeted refresh closes this gap. Recomputing features for the top 20\% of candidates under the stale ranking raises top-1,000 overlap to at least 0.971 in every setting, and $p=0.3$ recovers the subset exactly in all seven. Figure~\ref{fig:headline}(a) shows why: the curve rises steeply once $p$ passes the selection fraction $k/N=0.1$, as Proposition~\ref{prop:band} predicts, and saturates by $p=0.3$. We therefore set $p=0.3$ for a top-10\% selection, the selection fraction plus a margin of 0.2 for the drift.

\paragraph{Refresh allocation.}
The stratified subset recovers less than the top-$k$ subset, as Section~\ref{sec:method_theory} predicts: a global refresh set concentrates on the source that scores highest under the target, so the other source keeps its calibrated stale scores. Figure~\ref{fig:allocation_chain}(a) confirms this in both directions: global allocation misses Dolly examples under the \gsm target and \gsm examples under MMLU. Allocating the same budget per stratum removes the shortfall and recovers the stratified subset at 0.992 to 1.000 under \gsm and at 0.977 and 0.919 under MMLU.

\paragraph{Alternatives and calibration.}
Targeted refresh outperforms two simpler ways to spend less on gradients: reusing checkpoint-40 features at checkpoint 120, which halves the number of scoring passes, recovers 0.57 to 0.77 of the top-1,000 subset, and refreshing a random 30\% of the pool recovers 0.89 to 0.92 at the same budget, with lower stratified overlap and downstream accuracy (Appendix~\ref{app:alternatives}). The affine calibration changes post-warmup overlap at $p\ge0.2$ by at most 0.001 and adds 1.9 to 7.4 points at $p=0.1$ and on early or long intervals (Appendix~\ref{app:calibration}).

\subsection{Checkpoint distance, repeated selection, and the online check}
\label{sec:results_distance}

\begin{figure}[t]
\centering
\includegraphics[width=\textwidth]{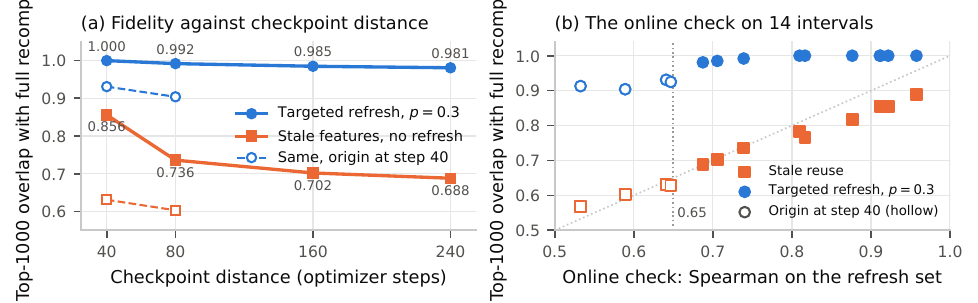}
\caption{(a)~Fidelity against checkpoint distance on Qwen3-4B seed 0, single refresh from the origin; filled markers start at checkpoint 80 and hollow markers at checkpoint 40. (b)~The online check (Spearman correlation between stale and recomputed scores on the $p=0.3$ refresh set) against the top-1,000 overlap of stale reuse and of targeted refresh on all fourteen single-refresh intervals (Appendix~\ref{app:check}).}
\label{fig:decay_check}
\end{figure}

\paragraph{Distance, origin, and repeated selection.}
Figure~\ref{fig:decay_check}(a) extends the single-refresh measurement to longer intervals and to an earlier origin. From checkpoint 80, stale fidelity decays from 0.856 at distance 40 to 0.688 at distance 240 while targeted refresh at $p=0.3$ stays between 1.000 and 0.981, and an origin at checkpoint 40 lowers both to 0.603 and 0.904 at distance 80, against 0.736 and 0.992 from checkpoint 80. The saving in the cost model of Section~\ref{sec:setup} accrues when the cache is iterated over the checkpoints that LESS scores, so Figure~\ref{fig:allocation_chain}(b) runs one full pass at checkpoint 80 followed by three consecutive refreshes at checkpoints 160, 240, and 320 of the warmup run, at a cumulative cost of 1.9 features per example against 4.0. Fidelity holds across the chain: global allocation recovers the top-1,000 subset at 0.979 or higher at every checkpoint and the checkpoint-averaged LESS score at 0.999, per-stratum allocation recovers the stratified subset at 0.939 or higher and at 0.990 for the aggregate, and stale features fall to 0.69 to 0.74. Chains that start at checkpoint 40, inside the unstable early interval, reach 0.90 to 1.00 top-1,000 and 0.62 to 0.84 stratified overlap at checkpoint 120 (Appendix~\ref{app:chain}, with a per-cohort calibration for the mixed-age cache).

\paragraph{The online check.}
Because the refresh computes exact scores for the refresh set anyway, the agreement between stale and recomputed scores on that set is available at every refresh, and Figure~\ref{fig:decay_check}(b) shows that it predicts fidelity. Across all fourteen single-refresh intervals it tracks the top-1,000 overlap of stale reuse to within 0.07, and the global rank correlation stays near 0.91 across the same intervals (Appendix~\ref{app:check}). The four intervals on which $p=0.3$ recovers less than 0.95 of the subset are exactly those with agreement below 0.65, every interval above 0.68 recovers at least 0.981, and in the chains the check flags the first step of the early-origin Qwen3 chains at 0.64. At the long single-refresh distances, where the agreement drops to 0.69 to 0.74 while targeted refresh still recovers 0.98, a threshold of 0.75 triggers extra gradient computation and keeps the selection intact.

\subsection{Downstream quality and training-run variance}
\label{sec:results_quality}

\begin{figure}[t]
\centering
\begin{subfigure}[t]{0.29\textwidth}
\centering
\includegraphics[width=\textwidth]{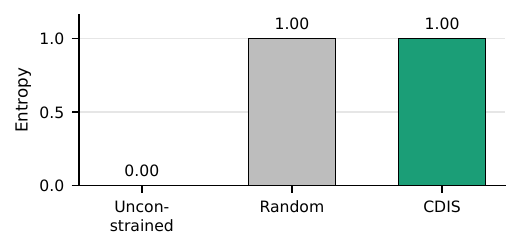}
\caption{Source entropy of the selected subset.}
\end{subfigure}\hspace{0.05\textwidth}
\begin{subfigure}[t]{0.29\textwidth}
\centering
\includegraphics[width=\textwidth]{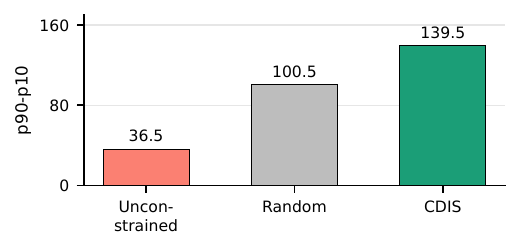}
\caption{Prompt-length coverage (p10 to p90).}
\end{subfigure}
\caption{Unconstrained top-$k$ selection on influence scores collapses to a single source and a narrow length window; \method keeps source balance and a wide length range (per-seed profiles in Appendix~\ref{app:profiles}).}
\label{fig:collapse}
\end{figure}

\begin{table}[t]
\centering
\scriptsize
\setlength{\tabcolsep}{2.8pt}
\begin{tabular}{lcccccc}
\toprule
& & \multicolumn{2}{c}{Full \gsm} & \multicolumn{2}{c}{Full \asdiv (transfer)} \\
\cmidrule(lr){3-4}\cmidrule(lr){5-6}
Method & Recompute & EM & \method $-$ method [95\% CI] & EM & \method $-$ method [95\% CI] \\
\midrule
Unconstrained top-$k$, full recompute & $N$ & 0.263 & $+$24.7 [$+$23.1, $+$26.3] & 0.163 & $+$15.6 [$+$14.4, $+$16.7] \\
Plain random & 0 & 0.389 & $+$12.2 [$+$10.6, $+$13.7] & 0.248 & $+$7.1 [$+$6.0, $+$8.1] \\
Length-matched random & 0 & 0.323 & $+$18.7 [$+$17.0, $+$20.3] & 0.170 & $+$14.9 [$+$13.7, $+$16.0] \\
Stratified, full recompute & $N$ & 0.553 & $-$4.3 [$-$5.5, $-$3.0] & 0.339 & $-$2.0 [$-$2.9, $-$1.1] \\
\oursrow
\method, $p=0.3$, global allocation & $0.3N$ & 0.510 & --- & 0.319 & --- \\
\bottomrule
\end{tabular}
\caption{Downstream quality on Qwen3-4B-Base (top-10\% of the 10k \gsm+Dolly pool, five-epoch LoRA, same adapters on full \gsm and full \asdiv). EM is the three-seed mean (per-seed values in Appendix~\ref{app:perseed}); differences are percentage points with paired, seed-stratified bootstrap 95\% CIs over per-item correctness (10,000 resamples).}
\label{tab:quality}
\end{table}

\paragraph{Unconstrained influence selection collapses.}
A faithful cache reproduces full recomputation, and full recomputation itself selects poorly when its scores are consumed without constraints. Top-$k$ selection on raw influence scores picks 1,000 \gsm examples and zero Dolly examples in all three seeds, concentrated in a 30 to 66 token prompt-length window (Figure~\ref{fig:collapse}), with the lowest training loss and the lowest test accuracy of every method we test: 0.263 on full \gsm against 0.389 for plain random (Table~\ref{tab:quality}). Because the collapse appears at full recomputation and the refresh preserves it exactly, the score signal itself produces it, and since length-matched random selection also falls below plain random in the three-seed mean, length concentration accounts for part of the loss and source concentration for the rest.

\paragraph{Stratified selection recovers a large margin, and the cache keeps most of it.}
Once the same scores pass through the stratified rule, the picture reverses. Stratified selection with full recomputation reaches 0.553 on full \gsm and 0.339 on \asdiv, 16.4 [14.8, 18.0] and 9.0 [8.0, 10.1] points above plain random, and \method with $p=0.3$ and global allocation follows at 0.510 and 0.319, which is 12.2 and 7.1 points above plain random and 24.7 and 15.6 above unconstrained selection. Both stratified selectors exceed both random controls on every seed and both test sets, and unconstrained selection stays below plain random on every seed on \asdiv and on two of three seeds on \gsm, within 0.001 on the third (Appendix~\ref{app:perseed}). The difference to full recomputation is 4.3 [3.0, 5.5] points on \gsm and 2.0 [1.1, 2.9] on \asdiv, from selections that agree with the reference subset in 0.86 to 0.94 of their examples.

\paragraph{Training-run variance.}
The paired intervals cover item sampling, so we put that difference on the scale of training noise by freezing four selections and retraining each with three training seeds (Appendix~\ref{app:noise}). Retraining an identical selection moves full \gsm exact match by up to 8.9 points, with a pooled within-selection standard deviation of 4.25 points; \method stays below full recomputation in every retraining pair, by 4.9 and 7.5 points in the retrain means, so its cost amounts to one to two standard deviations while its margin over random selection amounts to three. \method therefore lowers selection-cycle cost by 3.37$\times$ at a cost of one to two training-run standard deviations on the target task and two points on transfer, and keeps a margin over random selection several times the noise floor.

\subsection{Selection rules}
\label{sec:results_rules}

The stratified rule needs both of its constraints. On the seed-0 GSM200 probe (Appendix~\ref{app:rules}; sampling standard error 3.5 points), length quotas alone still select 1,000 \gsm examples and reach 0.100, source quotas alone balance the sources and reach 0.410, and the combination reaches 0.655, with \method at $p=0.3$ tracking each variant. The equal source split is also the best one, because raising the \gsm share to 70\% and 80\% lowers GSM200 accuracy to 0.400 and 0.350, differences far beyond the probe's sampling error, and BIDS-style normalization, the closest balanced-influence rule, is preserved as well (0.615 with full recomputation, 0.625 at $p=0.3$).

Source labels can come from the cache itself. Clustering the cached checkpoint-80 features with $k$-means ($k=8$) recovers the \gsm/Dolly split with 99.8\% purity, quotas proportional to cluster size select 517 \gsm and 483 Dolly examples, and the refresh preserves this rule at 0.924 stratified overlap, with every differing example a Dolly-side swap as Section~\ref{sec:method_theory} predicts. Downstream on seed 0, the label-free rule reaches 0.579 on full \gsm with full recomputation, above plain random at 0.529, and 0.468 with $p=0.3$, a difference inside the training-run range of Appendix~\ref{app:noise}.

\subsection{Cost accounting}
\label{sec:results_cost}

The train-gradient stage drops from 3,935.75 to 1,087.33 seconds on Qwen3-4B-Base (3.62$\times$) and from 1,579.72 to 448.60 seconds on Llama-3.2-1B (3.52$\times$), and since validation-gradient extraction (106.70\,s) and CPU scoring, calibration, and selection (6.83\,s) are shared, the selection cycle drops from 4,049.28 to 1,200.86 seconds (3.37$\times$) while downstream LoRA training is unchanged (Appendix~\ref{app:cost}). The factors exceed 3.33$\times$ because the refreshed candidates under a \gsm target are shorter than the pool average (Appendix~\ref{app:profiles}). Because the cache costs one full pass, the saving begins with the second selection cycle, at 1.54$\times$ for two cycles, 2.29$\times$ for five, and 3.37$\times$ in the limit, and the four-checkpoint chain of Figure~\ref{fig:allocation_chain}(b) realizes 2.1$\times$ on the train-gradient stage.

\section{Conclusion}
\label{sec:discussion}

Influence rankings in LESS-style selection drift slowly between post-training checkpoints, and the drift concentrates at the top of the ranking, where a targeted refresh absorbs it: a refresh fraction above the selection fraction recovers the recomputed subset exactly on three seeds, two model families, two pools, and two targets at 3.5 to 3.6 times lower gradient wall-clock, and sustains 98\% fidelity over three consecutive refreshes at half the gradient cost. Per-stratum allocation carries Proposition~\ref{prop:band} to stratified selection, and with that rule \method keeps the downstream quality of full recomputation within one to two training-run standard deviations at a fraction of its cost; downstream training uses the global allocation, and the per-stratum selections and the MMLU target are measured at the selection level.

\bibliographystyle{iclr2027_conference}
\bibliography{refs}

\begin{thebibliography}{36}
\providecommand{\natexlab}[1]{#1}
\providecommand{\url}[1]{\texttt{#1}}
\expandafter\ifx\csname urlstyle\endcsname\relax
  \providecommand{\doi}[1]{doi: #1}\else
  \providecommand{\doi}{doi: \begingroup \urlstyle{rm}\Url}\fi

\bibitem[Albalak et~al.(2024)Albalak, Elazar, Xie, Longpre, Lambert, Wang,
  Muennighoff, Hou, Pan, Jeong, Raffel, Chang, Hashimoto, and
  Wang]{albalak2024survey}
Alon Albalak, Yanai Elazar, Sang~Michael Xie, Shayne Longpre, Nathan Lambert,
  Xinyi Wang, Niklas Muennighoff, Bairu Hou, Liangming Pan, Haewon Jeong, Colin
  Raffel, Shiyu Chang, Tatsunori Hashimoto, and William~Yang Wang.
\newblock A survey on data selection for language models.
\newblock \emph{Transactions on Machine Learning Research}, 2024.
\newblock URL \url{https://dblp.org/rec/journals/tmlr/AlbalakEXLL0MHP24}.

\bibitem[Bukharin et~al.(2024)Bukharin, Li, Wang, Yang, Yin, Li, Zhang, Zhao,
  and Jiang]{bukharin2024qdit}
Alexander Bukharin, Shiyang Li, Zhengyang Wang, Jingfeng Yang, Bing Yin, Xian
  Li, Chao Zhang, Tuo Zhao, and Haoming Jiang.
\newblock Data diversity matters for robust instruction tuning.
\newblock In \emph{Findings of the Association for Computational Linguistics:
  EMNLP 2024}, pp.\  3411--3425, Miami, Florida, USA, November 2024.
  Association for Computational Linguistics.
\newblock \doi{10.18653/v1/2024.findings-emnlp.195}.
\newblock URL \url{https://aclanthology.org/2024.findings-emnlp.195/}.

\bibitem[Chen et~al.(2026)Chen, Qi, Ai, Sun, Qiu, Zou, and He]{chen2026iprox}
Sirui Chen, Yunzhe Qi, Mengting Ai, Yifan Sun, Ruizhong Qiu, Jiaru Zou, and
  Jingrui He.
\newblock Influence-preserving proxies for gradient-based data selection in
  {LLM} fine-tuning.
\newblock In \emph{The Fourteenth International Conference on Learning
  Representations}, 2026.
\newblock URL \url{https://openreview.net/forum?id=PDNpRLxDlI}.

\bibitem[Choe et~al.(2025)Choe, Ahn, Bae, Zhao, Chung, Pratapa, Neiswanger,
  Strubell, Mitamura, Schneider, Hovy, Grosse, and Xing]{choe2025logra}
Sang~Keun Choe, Hwijeen Ahn, Juhan Bae, Kewen Zhao, Youngseog Chung, Adithya
  Pratapa, Willie Neiswanger, Emma Strubell, Teruko Mitamura, Jeff~G.
  Schneider, Eduard~H. Hovy, Roger~Baker Grosse, and Eric~P. Xing.
\newblock What is your data worth to {GPT}? {LLM}-scale data valuation with
  influence functions.
\newblock In \emph{Advances in Neural Information Processing Systems},
  volume~38, pp.\  145944--145985. Curran Associates, Inc., 2025.
\newblock \doi{10.52202/085713-4883}.
\newblock URL
  \url{https://proceedings.neurips.cc/paper_files/paper/2025/hash/d6d26053b977f8c589669fd201615119-Abstract-Conference.html}.

\bibitem[Cobbe et~al.(2021)Cobbe, Kosaraju, Bavarian, Chen, Jun, Kaiser,
  Plappert, Tworek, Hilton, Nakano, Hesse, and Schulman]{cobbe2021gsm8k}
Karl Cobbe, Vineet Kosaraju, Mohammad Bavarian, Mark Chen, Heewoo Jun, Lukasz
  Kaiser, Matthias Plappert, Jerry Tworek, Jacob Hilton, Reiichiro Nakano,
  Christopher Hesse, and John Schulman.
\newblock Training verifiers to solve math word problems.
\newblock \emph{arXiv preprint arXiv:2110.14168}, 2021.
\newblock \doi{10.48550/arXiv.2110.14168}.
\newblock URL \url{https://arxiv.org/abs/2110.14168}.

\bibitem[Conover et~al.(2023)Conover, Hayes, Mathur, Xie, Wan, Shah, Ghodsi,
  Wendell, Zaharia, and Xin]{databricks2023dolly}
Mike Conover, Matt Hayes, Ankit Mathur, Jianwei Xie, Jun Wan, Sam Shah, Ali
  Ghodsi, Patrick Wendell, Matei Zaharia, and Reynold Xin.
\newblock Free dolly: Introducing the world's first truly open
  instruction-tuned {LLM}, April 2023.
\newblock URL
  \url{https://www.databricks.com/blog/2023/04/12/dolly-first-open-commercially-viable-instruction-tuned-llm}.
\newblock Databricks blog post.

\bibitem[Dai et~al.(2025)Dai, Zhang, Ma, and Peng]{dai2025bids}
Qirun Dai, Dylan Zhang, Jiaqi~W. Ma, and Hao Peng.
\newblock Improving influence-based instruction tuning data selection for
  balanced learning of diverse capabilities.
\newblock In \emph{Findings of the Association for Computational Linguistics:
  EMNLP 2025}, pp.\  7079--7102, Suzhou, China, November 2025. Association for
  Computational Linguistics.
\newblock \doi{10.18653/v1/2025.findings-emnlp.373}.
\newblock URL \url{https://aclanthology.org/2025.findings-emnlp.373/}.

\bibitem[Engstrom et~al.(2024)Engstrom, Feldmann, and Madry]{engstrom2024dsdm}
Logan Engstrom, Axel Feldmann, and Aleksander Madry.
\newblock {DsDm}: Model-aware dataset selection with datamodels.
\newblock In \emph{Proceedings of the 41st International Conference on Machine
  Learning}, Proceedings of Machine Learning Research. PMLR, 2024.
\newblock URL \url{https://proceedings.mlr.press/v235/engstrom24a.html}.

\bibitem[Grattafiori et~al.(2024)]{grattafiori2024llama3}
Aaron Grattafiori et~al.
\newblock The {Llama} 3 herd of models.
\newblock \emph{arXiv preprint arXiv:2407.21783}, 2024.
\newblock URL \url{https://arxiv.org/abs/2407.21783}.

\bibitem[Grosse et~al.(2023)Grosse, Bae, Anil, Elhage, Tamkin, Tajdini,
  Steiner, Li, Durmus, Perez, Hubinger, Luko{\v{s}}i{\=u}t{\.e}, Nguyen,
  Joseph, McCandlish, Kaplan, and Bowman]{grosse2023influence}
Roger Grosse, Juhan Bae, Cem Anil, Nelson Elhage, Alex Tamkin, Amirhossein
  Tajdini, Benoit Steiner, Dustin Li, Esin Durmus, Ethan Perez, Evan Hubinger,
  Kamil{\.e} Luko{\v{s}}i{\=u}t{\.e}, Karina Nguyen, Nicholas Joseph, Sam
  McCandlish, Jared Kaplan, and Samuel~R. Bowman.
\newblock Studying large language model generalization with influence
  functions.
\newblock \emph{arXiv preprint arXiv:2308.03296}, 2023.
\newblock URL \url{https://arxiv.org/abs/2308.03296}.

\bibitem[Hendrycks et~al.(2021)Hendrycks, Burns, Basart, Zou, Mazeika, Song,
  and Steinhardt]{hendrycks2021mmlu}
Dan Hendrycks, Collin Burns, Steven Basart, Andy Zou, Mantas Mazeika, Dawn
  Song, and Jacob Steinhardt.
\newblock Measuring massive multitask language understanding.
\newblock In \emph{International Conference on Learning Representations}, 2021.
\newblock URL \url{https://dblp.org/rec/conf/iclr/HendrycksBBZMSS21}.

\bibitem[Hu et~al.(2022)Hu, Shen, Wallis, Allen-Zhu, Li, Wang, Wang, and
  Chen]{hu2022lora}
Edward~J. Hu, Yelong Shen, Phillip Wallis, Zeyuan Allen-Zhu, Yuanzhi Li, Shean
  Wang, Lu~Wang, and Weizhu Chen.
\newblock {LoRA}: Low-rank adaptation of large language models.
\newblock In \emph{International Conference on Learning Representations}, 2022.
\newblock URL \url{https://openreview.net/forum?id=nZeVKeeFYf9}.

\bibitem[Ilyas et~al.(2022)Ilyas, Park, Engstrom, Leclerc, and
  Madry]{ilyas2022datamodels}
Andrew Ilyas, Sung~Min Park, Logan Engstrom, Guillaume Leclerc, and Aleksander
  Madry.
\newblock Datamodels: Understanding predictions with data and data with
  predictions.
\newblock In \emph{Proceedings of the 39th International Conference on Machine
  Learning}, volume 162 of \emph{Proceedings of Machine Learning Research},
  pp.\  9525--9587. PMLR, 2022.
\newblock URL \url{https://proceedings.mlr.press/v162/ilyas22a.html}.

\bibitem[Jiang et~al.(2023)Jiang, Sablayrolles, Mensch, Bamford, Chaplot,
  de~las Casas, Bressand, Lengyel, Lample, Saulnier, Lavaud, Lachaux, Stock,
  Le~Scao, Lavril, Wang, Lacroix, and El~Sayed]{jiang2023mistral}
Albert~Q. Jiang, Alexandre Sablayrolles, Arthur Mensch, Chris Bamford,
  Devendra~Singh Chaplot, Diego de~las Casas, Florian Bressand, Gianna Lengyel,
  Guillaume Lample, Lucile Saulnier, L{\'e}lio~Renard Lavaud, Marie-Anne
  Lachaux, Pierre Stock, Teven Le~Scao, Thibaut Lavril, Thomas Wang,
  Timoth{\'e}e Lacroix, and William El~Sayed.
\newblock {Mistral} 7b.
\newblock \emph{arXiv preprint arXiv:2310.06825}, 2023.
\newblock URL \url{https://arxiv.org/abs/2310.06825}.

\bibitem[Killamsetty et~al.(2021)Killamsetty, Sivasubramanian, Ramakrishnan,
  De, and Iyer]{killamsetty2021gradmatch}
Krishnateja Killamsetty, Durga Sivasubramanian, Ganesh Ramakrishnan, Abir De,
  and Rishabh Iyer.
\newblock {GRAD-MATCH}: Gradient matching based data subset selection for
  efficient deep model training.
\newblock In \emph{Proceedings of the 38th International Conference on Machine
  Learning}, Proceedings of Machine Learning Research. PMLR, 2021.
\newblock URL \url{https://proceedings.mlr.press/v139/killamsetty21a.html}.

\bibitem[Koh \& Liang(2017)Koh and Liang]{koh2017influence}
Pang~Wei Koh and Percy Liang.
\newblock Understanding black-box predictions via influence functions.
\newblock In \emph{Proceedings of the 34th International Conference on Machine
  Learning}, volume~70 of \emph{Proceedings of Machine Learning Research}, pp.\
   1885--1894. PMLR, 06--11 Aug 2017.
\newblock URL \url{https://proceedings.mlr.press/v70/koh17a.html}.

\bibitem[Kwon et~al.(2024)Kwon, Wu, Wu, and Zou]{kwon2024datainf}
Yongchan Kwon, Eric Wu, Kevin Wu, and James Zou.
\newblock {DataInf}: Efficiently estimating data influence in {LoRA}-tuned
  {LLMs} and diffusion models.
\newblock In \emph{The Twelfth International Conference on Learning
  Representations}, 2024.
\newblock URL \url{https://openreview.net/forum?id=9m02ib92Wz}.

\bibitem[Liu et~al.(2024)Liu, Zeng, He, Jiang, and He]{liu2024deita}
Wei Liu, Weihao Zeng, Keqing He, Yong Jiang, and Junxian He.
\newblock What makes good data for alignment? a comprehensive study of
  automatic data selection in instruction tuning.
\newblock In \emph{The Twelfth International Conference on Learning
  Representations}, 2024.
\newblock URL \url{https://openreview.net/forum?id=BTKAeLqLMw}.

\bibitem[Longpre et~al.(2023)Longpre, Hou, Vu, Webson, Chung, Tay, Zhou, Le,
  Zoph, Wei, and Roberts]{longpre2023flan}
Shayne Longpre, Le~Hou, Tu~Vu, Albert Webson, Hyung~Won Chung, Yi~Tay, Denny
  Zhou, Quoc~V. Le, Barret Zoph, Jason Wei, and Adam Roberts.
\newblock The {Flan} collection: Designing data and methods for effective
  instruction tuning.
\newblock In \emph{Proceedings of the 40th International Conference on Machine
  Learning}, Proceedings of Machine Learning Research. PMLR, 2023.
\newblock URL \url{https://dblp.org/rec/conf/icml/LongpreHVWCTZLZ23}.

\bibitem[Lu et~al.(2024)Lu, Yuan, Yuan, Lin, Lin, Tan, Zhou, and
  Zhou]{lu2024instag}
Keming Lu, Hongyi Yuan, Zheng Yuan, Runji Lin, Junyang Lin, Chuanqi Tan, Chang
  Zhou, and Jingren Zhou.
\newblock {{\#}InsTag}: Instruction tagging for analyzing supervised
  fine-tuning of large language models.
\newblock In \emph{The Twelfth International Conference on Learning
  Representations}, 2024.
\newblock URL \url{https://openreview.net/forum?id=pszewhybU9}.

\bibitem[{Meta}(2024)]{meta2024llama32}
{Meta}.
\newblock {Llama-3.2-1B} model card, 2024.
\newblock URL \url{https://huggingface.co/meta-llama/Llama-3.2-1B}.
\newblock Official model card; accessed September 25, 2026.

\bibitem[Miao et~al.(2020)Miao, Liang, and Su]{miao2020asdiv}
Shen-yun Miao, Chao-Chun Liang, and Keh-Yih Su.
\newblock A diverse corpus for evaluating and developing english math word
  problem solvers.
\newblock In \emph{Proceedings of the 58th Annual Meeting of the Association
  for Computational Linguistics}, pp.\  975--984, Online, 2020. Association for
  Computational Linguistics.
\newblock \doi{10.18653/v1/2020.acl-main.92}.
\newblock URL \url{https://aclanthology.org/2020.acl-main.92/}.

\bibitem[Mirzasoleiman et~al.(2020)Mirzasoleiman, Bilmes, and
  Leskovec]{mirzasoleiman2020craig}
Baharan Mirzasoleiman, Jeff Bilmes, and Jure Leskovec.
\newblock Coresets for data-efficient training of machine learning models.
\newblock In \emph{Proceedings of the 37th International Conference on Machine
  Learning}, Proceedings of Machine Learning Research. PMLR, 2020.
\newblock URL \url{https://dblp.org/rec/conf/icml/MirzasoleimanBL20}.

\bibitem[{Mistral AI}(2024)]{mistral2024v03}
{Mistral AI}.
\newblock {Mistral-7B-v0.3} model card, 2024.
\newblock URL \url{https://huggingface.co/mistralai/Mistral-7B-v0.3}.
\newblock Official model card; accessed September 25, 2026.

\bibitem[Pan et~al.(2024)Pan, Huang, Kang, Liu, Lu, and Cheng]{pan2024gdig}
Xingyuan Pan, Luyang Huang, Liyan Kang, Zhicheng Liu, Yu~Lu, and Shanbo Cheng.
\newblock {G}-{DIG}: Towards gradient-based {DI}verse and hi{G}h-quality
  instruction data selection for machine translation.
\newblock In \emph{Proceedings of the 62nd Annual Meeting of the Association
  for Computational Linguistics (Volume 1: Long Papers)}, pp.\  15395--15406,
  Bangkok, Thailand, August 2024. Association for Computational Linguistics.
\newblock \doi{10.18653/v1/2024.acl-long.821}.
\newblock URL \url{https://aclanthology.org/2024.acl-long.821/}.

\bibitem[Park et~al.(2023)Park, Georgiev, Ilyas, Leclerc, and
  Madry]{park2023trak}
Sung~Min Park, Kristian Georgiev, Andrew Ilyas, Guillaume Leclerc, and
  Aleksander Madry.
\newblock {TRAK}: Attributing model behavior at scale.
\newblock In \emph{Proceedings of the 40th International Conference on Machine
  Learning}, volume 202 of \emph{Proceedings of Machine Learning Research},
  pp.\  27074--27113. PMLR, 23--29 Jul 2023.
\newblock URL \url{https://proceedings.mlr.press/v202/park23c.html}.

\bibitem[Pruthi et~al.(2020)Pruthi, Liu, Kale, and
  Sundararajan]{pruthi2020tracin}
Garima Pruthi, Frederick Liu, Satyen Kale, and Mukund Sundararajan.
\newblock Estimating training data influence by tracing gradient descent.
\newblock In \emph{Advances in Neural Information Processing Systems},
  volume~33, pp.\  19920--19930, 2020.
\newblock URL
  \url{https://proceedings.neurips.cc/paper/2020/hash/e6385d39ec9394f2f3a354d9d2b88eec-Abstract.html}.

\bibitem[Wang et~al.(2024)Wang, Wu, Song, Mittal, and Jia]{wang2024greats}
Jiachen~T. Wang, Tong Wu, Dawn Song, Prateek Mittal, and Ruoxi Jia.
\newblock {GREATS}: Online selection of high-quality data for {LLM} training in
  every iteration.
\newblock In \emph{Advances in Neural Information Processing Systems 37
  (NeurIPS 2024)}, 2024.
\newblock URL \url{https://openreview.net/forum?id=232VcN8tSx}.

\bibitem[Wang et~al.(2025)Wang, Song, Zou, Mittal, and Jia]{wang2025temporal}
Jiachen~T. Wang, Dawn Song, James Zou, Prateek Mittal, and Ruoxi Jia.
\newblock Capturing the temporal dependence of training data influence.
\newblock In \emph{The Thirteenth International Conference on Learning
  Representations}, 2025.
\newblock URL \url{https://openreview.net/forum?id=uHLgDEgiS5}.

\bibitem[Xia et~al.(2024)Xia, Malladi, Gururangan, Arora, and
  Chen]{xia2024less}
Mengzhou Xia, Sadhika Malladi, Suchin Gururangan, Sanjeev Arora, and Danqi
  Chen.
\newblock {LESS}: Selecting influential data for targeted instruction tuning.
\newblock In \emph{Proceedings of the 41st International Conference on Machine
  Learning}, volume 235 of \emph{Proceedings of Machine Learning Research},
  pp.\  54104--54132. PMLR, 21--27 Jul 2024.
\newblock URL \url{https://proceedings.mlr.press/v235/xia24c.html}.

\bibitem[Xia et~al.(2025)Xia, Yu, Dang, Yang, Wu, Tian, Chang, and
  Lin]{xia2025random}
Tingyu Xia, Bowen Yu, Kai Dang, An~Yang, Yuan Wu, Yuan Tian, Yi~Chang, and
  Junyang Lin.
\newblock Rethinking data selection at scale: Random selection is almost all
  you need.
\newblock In \emph{Findings of the Association for Computational Linguistics:
  EMNLP 2025}, pp.\  2698--2711, Suzhou, China, November 2025. Association for
  Computational Linguistics.
\newblock \doi{10.18653/v1/2025.findings-emnlp.146}.
\newblock URL \url{https://aclanthology.org/2025.findings-emnlp.146/}.

\bibitem[Yang et~al.(2025)Yang, Li, Yang, Zhang, Hui, Zheng, Yu, Gao, Huang,
  Lv, Zheng, Liu, Zhou, Huang, Hu, Ge, Wei, Lin, Tang, Yang, Tu, Zhang, Yang,
  Yang, Zhou, Zhou, Lin, Dang, Bao, Yang, Yu, Deng, Li, Xue, Li, Zhang, Wang,
  Zhu, Men, Gao, Liu, Luo, Li, Tang, Yin, Ren, Wang, Zhang, Ren, Fan, Su,
  Zhang, Zhang, Wan, Liu, Wang, Cui, Zhang, Zhou, and Qiu]{yang2025qwen3}
An~Yang, Anfeng Li, Baosong Yang, Beichen Zhang, Binyuan Hui, Bo~Zheng, Bowen
  Yu, Chang Gao, Chengen Huang, Chenxu Lv, Chujie Zheng, Dayiheng Liu, Fan
  Zhou, Fei Huang, Feng Hu, Hao Ge, Haoran Wei, Huan Lin, Jialong Tang, Jian
  Yang, Jianhong Tu, Jianwei Zhang, Jianxin Yang, Jiaxi Yang, Jing Zhou,
  Jingren Zhou, Junyang Lin, Kai Dang, Keqin Bao, Kexin Yang, Le~Yu, Lianghao
  Deng, Mei Li, Mingfeng Xue, Mingze Li, Pei Zhang, Peng Wang, Qin Zhu, Rui
  Men, Ruize Gao, Shixuan Liu, Shuang Luo, Tianhao Li, Tianyi Tang, Wenbiao
  Yin, Xingzhang Ren, Xinyu Wang, Xinyu Zhang, Xuancheng Ren, Yang Fan, Yang
  Su, Yichang Zhang, Yinger Zhang, Yu~Wan, Yuqiong Liu, Zekun Wang, Zeyu Cui,
  Zhenru Zhang, Zhipeng Zhou, and Zihan Qiu.
\newblock {Qwen3} technical report.
\newblock \emph{arXiv preprint arXiv:2505.09388}, 2025.
\newblock \doi{10.48550/arXiv.2505.09388}.
\newblock URL \url{https://arxiv.org/abs/2505.09388}.

\bibitem[Yu et~al.(2024)Yu, Das, and Xiong]{yu2024mates}
Zichun Yu, Spandan Das, and Chenyan Xiong.
\newblock {MATES}: Model-aware data selection for efficient pretraining with
  data influence models.
\newblock In \emph{Advances in Neural Information Processing Systems},
  volume~37, 2024.
\newblock URL \url{https://dblp.org/rec/conf/nips/YuDX24}.

\bibitem[Zhang et~al.(2025{\natexlab{a}})Zhang, Zhang, Liu, Jin, Yang, Zheng,
  Liu, and Guo]{zhang2025d3}
Jia Zhang, Chen-Xi Zhang, Yao Liu, Yi-Xuan Jin, Xiao-Wen Yang, Bo~Zheng,
  Yi~Liu, and Lan-Zhe Guo.
\newblock {D3}: Diversity, difficulty, and dependability-aware data selection
  for sample-efficient {LLM} instruction tuning.
\newblock In \emph{Proceedings of the Thirty-Fourth International Joint
  Conference on Artificial Intelligence, {IJCAI-25}}, pp.\  8348--8356.
  International Joint Conferences on Artificial Intelligence Organization,
  2025{\natexlab{a}}.
\newblock \doi{10.24963/ijcai.2025/928}.
\newblock URL \url{https://www.ijcai.org/proceedings/2025/928}.

\bibitem[Zhang et~al.(2025{\natexlab{b}})Zhang, Qin, Pi, Zhang, Pan, and
  Zhang]{zhang2025tagcos}
Jipeng Zhang, Yaxuan Qin, Renjie Pi, Weizhong Zhang, Rui Pan, and Tong Zhang.
\newblock {TAGCOS}: Task-agnostic gradient clustered {CO}reset selection for
  instruction tuning data.
\newblock In \emph{Findings of the Association for Computational Linguistics:
  NAACL 2025}, pp.\  4686--4701, Albuquerque, New Mexico, April
  2025{\natexlab{b}}. Association for Computational Linguistics.
\newblock \doi{10.18653/v1/2025.findings-naacl.264}.
\newblock URL \url{https://aclanthology.org/2025.findings-naacl.264/}.

\bibitem[Zhou et~al.(2023)Zhou, Liu, Xu, Iyer, Sun, Mao, Ma, Efrat, Yu, Yu,
  Zhang, Ghosh, Lewis, Zettlemoyer, and Levy]{zhou2023lima}
Chunting Zhou, Pengfei Liu, Puxin Xu, Srinivasan Iyer, Jiao Sun, Yuning Mao,
  Xuezhe Ma, Avia Efrat, Ping Yu, Lili Yu, Susan Zhang, Gargi Ghosh, Mike
  Lewis, Luke Zettlemoyer, and Omer Levy.
\newblock {LIMA}: Less is more for alignment.
\newblock In \emph{Advances in Neural Information Processing Systems},
  volume~36, 2023.
\newblock URL \url{https://openreview.net/forum?id=KBMOKmX2he}.

\end{thebibliography}

\clearpage
\appendix
\section{Proof of Proposition~\ref{prop:band}}
\label{app:proof}

Let $\mathcal{M}=\{i: c(\tilde{s}_i)\ge\tau-\varepsilon\}$ and assume $\mathcal{R}\supseteq\mathcal{M}$. Consider first any $i\in\mathcal{T}_k$. By definition $s_i^r\ge\tau$, and by the error bound $c(\tilde{s}_i)\ge s_i^r-\varepsilon\ge\tau-\varepsilon$, so $i\in\mathcal{M}\subseteq\mathcal{R}$ and the hybrid score is exact, $\hat{s}_i=s_i^r\ge\tau$. Consider next any $j\notin\mathcal{R}$. Then $j\notin\mathcal{M}$, so $c(\tilde{s}_j)<\tau-\varepsilon$, and the error bound gives $s_j^r\le c(\tilde{s}_j)+\varepsilon<\tau$, hence $j\notin\mathcal{T}_k$; its hybrid score is $\hat{s}_j=c(\tilde{s}_j)<\tau-\varepsilon<\tau$. Finally, any $j\in\mathcal{R}\setminus\mathcal{T}_k$ has exact hybrid score $\hat{s}_j=s_j^r<\tau$ because the scores are distinct and $j$ lies outside the top $k$. Hence every member of $\mathcal{T}_k$ has hybrid score at least $\tau$ and every non-member has hybrid score below $\tau$, so the top-$k$ subset under $\hat{s}$ is $\mathcal{T}_k$. \qed

When $\mathcal{R}$ is the top-$pN$ set under $\tilde{s}$ and the calibration slope $a$ is positive, $\mathcal{R}$ is also the top-$pN$ set under $c(\tilde{s})$, so the condition $\mathcal{R}\supseteq\mathcal{M}$ holds exactly when $pN\ge|\mathcal{M}|$. Since $\mathcal{T}_k\subseteq\mathcal{M}$, this requires $p\ge k/N$; the excess $|\mathcal{M}|-k$ counts the candidates outside $\mathcal{T}_k$ whose calibrated stale score lies within $\varepsilon$ below the threshold, together with the candidates whose stale score exceeds the threshold although their exact score does not. For a stratified rule with strata $(c,b)$, the argument applies within each stratum with its own threshold $\tau_{c,b}$ and quota $k_{c,b}$, and exact recovery of the whole subset requires $\mathcal{R}$ to contain the union of the stratum-level sets $\mathcal{M}_{c,b}$.

\section{Algorithm}
\label{app:algorithm}

\begin{algorithm}[H]
\caption{\method: targeted partial refresh with stratified selection}
\label{alg:cdis}
\begin{algorithmic}[1]
\Require Pool $\mathcal{D}=\{x_i\}_{i=1}^N$; cached features $G^\ell=\{g_i^\ell\}_{i=1}^N$; validation direction $v^r$; refresh fraction $p$; subset size $k$; stratum assignment $(c_i,b_i)$; stratum quotas $\{k_{c,b}\}$.
\Ensure Selected subset $\mathcal{S}$, updated cache, and check statistic $\rho$.
\State $\tilde{s}_i \gets \langle g_i^\ell, v^r\rangle$ for all $i\in[N]$ \Comment{stale scores}
\State $\mathcal{R}_p \gets \TopK_{\lceil pN\rceil}([N]; \tilde{s})$ \Comment{refresh set}
\For{$i \in \mathcal{R}_p$}
    \State recompute $g_i^r$ at $\theta_r$; $s_i^r \gets \langle g_i^r, v^r\rangle$; $g_i^\ell \gets g_i^r$ \Comment{update cache}
\EndFor
\State $(a,b) \gets \arg\min_{a,b}\sum_{i\in\mathcal{R}_p}(a\tilde{s}_i+b-s_i^r)^2$ \Comment{affine calibration}
\State $\rho \gets \mathrm{Spearman}(\{\tilde{s}_i\}_{i\in\mathcal{R}_p},\{s_i^r\}_{i\in\mathcal{R}_p})$ \Comment{online check, Section~\ref{sec:method_check}}
\State $\hat{s}_i \gets s_i^r$ if $i\in\mathcal{R}_p$ else $a\tilde{s}_i+b$, for all $i\in[N]$
\State $\mathcal{S}\gets \emptyset$
\ForAll{strata $(c,b)$}
    \State $\mathcal{S}\gets \mathcal{S}\cup \TopK_{k_{c,b}}(\{i: c_i=c,\ b_i=b\}; \hat{s})$
\EndFor
\If{$|\mathcal{S}|<k$}
    \State $\mathcal{S}\gets\mathcal{S}\cup \TopK_{k-|\mathcal{S}|}([N]\setminus\mathcal{S};\hat{s})$ \Comment{top up underfilled strata}
\EndIf
\State \Return $\mathcal{S}$, $G^\ell$, $\rho$
\end{algorithmic}
\end{algorithm}

\section{Implementation details}
\label{app:impl}

\paragraph{Candidate pool construction.}
For each selection seed we sample 5,000 \gsm training examples and 5,000 Dolly examples without replacement, convert the \gsm examples into a chat format with a user turn that asks the model to solve the problem and an assistant turn with the reference solution, keep the Dolly examples in their processed instruction format, and permute the concatenation with the same seed. Each example carries an index, a source label, and a prompt-length proxy equal to the whitespace-token count of all message contents. The \gsm+FLAN-v2 pool replaces the Dolly half by 5,000 FLAN-v2 examples with the same procedure.

\paragraph{Warmup and gradient stores.}
The warmup follows LESS: LoRA with rank 8, $\alpha=32$, dropout 0.1 on the query, key, value, and output projections; a random 7\% of Dolly; one epoch; per-device batch size 1 with gradient accumulation 8; learning rate $2\cdot10^{-5}$ with a linear schedule, warmup ratio 0.03, and weight decay 0; checkpoints every 40 steps. Train features are Adam-aware projected gradients with $d=8{,}192$ from the LESS implementation, computed for the whole pool at checkpoints 40, 80, and 120. Validation gradients are computed at each checkpoint from the 100 \gsm validation examples of the LESS pipeline, for both model families, and averaged into $v^t$. The three-epoch warmup run uses the same data, schedule type, and checkpoint spacing and stops after 396 steps.

\paragraph{Refresh and selection.}
For a refresh fraction $p$, the refresh set is the top $\lceil pN\rceil$ candidates by stale score, the affine calibration is a global least-squares fit on the refresh set, and the hybrid scores follow Eq.~\ref{eq:calib}. The stratified rule allocates 500 examples to each source, splits each source into four equal-count length buckets by the prompt-length proxy, divides each source quota evenly across buckets with largest-remainder rounding, fills each stratum by descending hybrid score, and tops up underfilled strata from the same source and then from the whole pool. The BIDS-style rule standardizes scores within each source, or within each source and length bucket, and takes a global top-$k$. The label-free rule runs $k$-means with $k=8$ on the checkpoint-80 features and allocates quotas in proportion to cluster size, or uniformly across clusters for the ablation.

\paragraph{Downstream training and evaluation.}
Each selected subset trains a LoRA adapter for five epochs with the warmup hyperparameters and maximum sequence length 512. Evaluation uses the LESS 8-shot \gsm evaluator with at most 256 new tokens and batch size 4 on the full \gsm test set and on the prepared \asdiv split. Exact match lowercases prediction and answer, removes punctuation, and collapses whitespace before comparison. Paired bootstrap intervals resample items within each selection seed 10,000 times and report the 2.5th and 97.5th percentiles of the mean per-item difference. Table~\ref{tab:settings} lists the settings.

\begin{table}[H]
\centering
\footnotesize
\begin{tabular}{lL{0.62\textwidth}}
\toprule
Component & Setting \\
\midrule
Main model; second family & Qwen3-4B-Base; Llama-3.2-1B \\
Candidate pool & 5k \gsm train + 5k Dolly (or 5k FLAN-v2), sampled per seed \\
Selection size & top 10\% = 1,000 examples \\
Projection dimension; gradient type & 8,192; Adam-aware projected gradients (LESS) \\
Cached and target checkpoints & 80 and 120 (main); 40$\to$80, 40$\to$120 (early); 80$\to$160/240/320 (three-epoch run) \\
Refresh fractions & 0, 0.05, 0.1, 0.2, 0.3, 0.5, 1 \\
Stratified rule & equal source quotas, four length buckets per source \\
LoRA & $r=8$, $\alpha=32$, dropout 0.1, modules q/k/v/o \\
Training & 5 epochs, max length 512, lr $2\cdot10^{-5}$, batch 1 $\times$ accumulation 8, AdamW, linear schedule \\
Evaluation & 8-shot GSM-style evaluator, 256 new tokens, exact match \\
Uncertainty & paired seed-stratified bootstrap, 10,000 resamples; 12-run training-seed decomposition \\
\bottomrule
\end{tabular}
\caption{Settings for the headline experiments.}
\label{tab:settings}
\end{table}

\section{Prompt and answer formats}
\label{app:prompts}

\gsm candidate examples use the user message \texttt{Solve the following math problem. Question: [problem] Answer:} and the reference solution as the assistant message. Dolly and FLAN-v2 examples keep their processed instruction-tuning messages with a normalized source label. The full \gsm test split is converted into the evaluator format with \texttt{question} and \texttt{answer} fields. For \asdiv we concatenate the problem body and question of the \texttt{EleutherAI/asdiv} validation split, extract a single numeric answer with a regular expression, skip the 73 items with fractional or non-numeric targets, and write answers with the GSM-style \texttt{\#\#\#\#} suffix so that the same evaluator, prompts, and decoding settings apply.

\section{Per-seed downstream results}
\label{app:perseed}

\begin{table}[H]
\centering
\footnotesize
\setlength{\tabcolsep}{4pt}
\begin{tabular}{lcccccccc}
\toprule
& \multicolumn{4}{c}{Full \gsm} & \multicolumn{4}{c}{Full \asdiv} \\
\cmidrule(lr){2-5}\cmidrule(lr){6-9}
Method & Seed 0 & Seed 1 & Seed 2 & Mean & Seed 0 & Seed 1 & Seed 2 & Mean \\
\midrule
Unconstrained top-$k$, full recompute & 0.196 & 0.237 & 0.356 & 0.263 & 0.126 & 0.164 & 0.199 & 0.163 \\
Plain random & 0.529 & 0.282 & 0.355 & 0.389 & 0.365 & 0.174 & 0.206 & 0.248 \\
Length-matched random & 0.356 & 0.434 & 0.180 & 0.323 & 0.182 & 0.247 & 0.081 & 0.170 \\
Stratified, full recompute & 0.661 & 0.453 & 0.544 & 0.553 & 0.467 & 0.250 & 0.299 & 0.339 \\
\method, $p=0.3$ & 0.616 & 0.477 & 0.438 & 0.510 & 0.436 & 0.289 & 0.232 & 0.319 \\
\bottomrule
\end{tabular}
\caption{Per-seed exact match behind Table~\ref{tab:quality}. Both stratified selectors exceed both random controls on every seed and both test sets; \method exceeds full recomputation on seed 1 and trails it on seeds 0 and 2, and the two random controls swap places across seeds.}
\label{tab:perseed}
\end{table}

\section{Selected-subset profiles}
\label{app:profiles}

\begin{table}[H]
\centering
\footnotesize
\setlength{\tabcolsep}{4pt}
\begin{tabular}{lcccccc}
\toprule
Method & Seed & \gsm & Dolly & Source entropy & Length mean & Length p10--p90 \\
\midrule
Unconstrained top-$k$ & 0 & 1000 & 0 & 0.000 & 44.8 & 30--64 \\
Unconstrained top-$k$ & 1 & 1000 & 0 & 0.000 & 46.9 & 30--69 \\
Plain random & 0 & 525 & 475 & 0.998 & 57.9 & 9--99 \\
Plain random & 1 & 490 & 510 & 1.000 & 61.0 & 10--121 \\
Length-matched random & 0 & 772 & 228 & 0.775 & 43.4 & 26--63 \\
Length-matched random & 1 & 844 & 156 & 0.625 & 46.0 & 29--69 \\
Stratified, full recompute & 0 & 500 & 500 & 1.000 & 69.2 & 10--155 \\
Stratified, full recompute & 1 & 500 & 500 & 1.000 & 54.2 & 9--105 \\
\method, $p=0.3$ & 0 & 500 & 500 & 1.000 & 67.0 & 10--154 \\
\method, $p=0.3$ & 1 & 500 & 500 & 1.000 & 66.3 & 10--145 \\
\bottomrule
\end{tabular}
\caption{Source counts, normalized source entropy, and prompt-length statistics of the selected subsets on seeds 0 and 1. Unconstrained selection collapses to one source and a narrow length window; the stratified rules keep source balance and a wide length range.}
\label{tab:profiles}
\end{table}

\section{Training-seed decomposition}
\label{app:noise}

\begin{table}[H]
\centering
\footnotesize
\begin{tabular}{lcccccc}
\toprule
Fixed selection & Seed a & Seed b & Seed c & Mean & SD & Range \\
\midrule
Seed-0 pool, stratified full recompute & 0.661 & 0.666 & 0.577 & 0.635 & 0.050 & 0.089 \\
Seed-0 pool, \method $p=0.3$ & 0.616 & 0.583 & 0.557 & 0.586 & 0.030 & 0.059 \\
Seed-2 pool, stratified full recompute & 0.544 & 0.567 & 0.479 & 0.530 & 0.046 & 0.088 \\
Seed-2 pool, \method $p=0.3$ & 0.438 & 0.503 & 0.425 & 0.455 & 0.042 & 0.077 \\
\bottomrule
\end{tabular}
\caption{Training-seed decomposition on full \gsm. Each row retrains the same selected subset with three training seeds (the original seed and two others) under identical hyperparameters. The pooled within-selection standard deviation is 0.0425.}
\label{tab:noise}
\end{table}

\section{Alternatives at equal or larger savings}
\label{app:alternatives}

\begin{table}[H]
\centering
\footnotesize
\setlength{\tabcolsep}{4pt}
\begin{tabular}{llccc}
\toprule
& & \multicolumn{3}{c}{Top-1,000 overlap at step 120} \\
\cmidrule(lr){3-5}
Strategy & Gradient cost & Qwen3 s0 & Qwen3 s1 & Llama \\
\midrule
Checkpoint skipping (reuse step-40 features) & half the scoring passes & 0.603 & 0.568 & 0.765 \\
Random refresh, 30\% & $0.3N$ per checkpoint & 0.890 & --- & 0.920 \\
Targeted refresh, 30\% & $0.3N$ per checkpoint & 1.000 & 1.000 & 1.000 \\
\bottomrule
\end{tabular}
\caption{Alternatives at equal or larger savings. Checkpoint skipping reuses features from twice the distance without refresh. Random and targeted refresh recompute the same number of features and differ in which candidates receive them. Random refresh on seed 1 was measured under the stratified rule (Table~\ref{tab:random_refresh}) and is omitted at the score level.}
\label{tab:alternatives}
\end{table}

\section{Random refresh on downstream quality}
\label{app:random_refresh}

\begin{table}[H]
\centering
\small
\begin{tabular}{llcccc}
\toprule
Seed & Refresh policy at 30\% & Stratified overlap & Top-500 prefix overlap & GSM200 & Full \gsm \\
\midrule
0 & Targeted & 0.927 & 1.000 & 0.600 & 0.616 \\
0 & Random   & 0.839 & 0.826 & 0.550 & 0.545 \\
1 & Targeted & 0.857 & 1.000 & 0.535 & 0.477 \\
1 & Random   & 0.739 & 0.734 & 0.310 & 0.322 \\
\bottomrule
\end{tabular}
\caption{Targeted against random refresh at the same 30\% budget under the stratified rule, with downstream exact match of the resulting adapters.}
\label{tab:random_refresh}
\end{table}

\section{Refresh-budget curve and calibration ablation}
\label{app:calibration}

Table~\ref{tab:budget_full} lists top-1,000 overlap with full recomputation for every setting, interval, and refresh fraction, with and without the affine calibration. Calibration changes post-warmup intervals at $p\ge0.2$ by at most 0.001, raises overlap by 1.9 to 7.4 points at $p=0.1$ and on early or long intervals, and the calibrated overlap is at least the uncalibrated one in every case.

\begin{table}[H]
\centering
\footnotesize
\setlength{\tabcolsep}{3.5pt}
\begin{tabular}{llccccccccc}
\toprule
& & stale & \multicolumn{2}{c}{$p=0.05$} & \multicolumn{2}{c}{$p=0.1$} & $p=0.2$ & $p=0.3$ & $p=0.5$ \\
\cmidrule(lr){4-5}\cmidrule(lr){6-7}
Setting & Interval & & cal. & raw & cal. & raw & & & \\
\midrule
Qwen3-4B s0 & 40$\to$80  & 0.631 & 0.648 & 0.631 & 0.689 & 0.631 & 0.826 & 0.931 & 0.999 \\
Qwen3-4B s0 & 40$\to$120 & 0.603 & 0.625 & 0.603 & 0.661 & 0.603 & 0.795 & 0.904 & 0.995 \\
Qwen3-4B s0 & 80$\to$120 & 0.856 & 0.859 & 0.859 & 0.894 & 0.894 & 0.999 & 1.000 & 1.000 \\
Qwen3-4B s1 & 40$\to$80  & 0.629 & 0.654 & 0.629 & 0.693 & 0.629 & 0.819 & 0.925 & 0.999 \\
Qwen3-4B s1 & 40$\to$120 & 0.568 & 0.599 & 0.568 & 0.643 & 0.569 & 0.785 & 0.913 & 0.998 \\
Qwen3-4B s1 & 80$\to$120 & 0.782 & 0.787 & 0.782 & 0.829 & 0.782 & 0.971 & 1.000 & 1.000 \\
Qwen3-4B s2 & 80$\to$120 & 0.817 & 0.824 & 0.818 & 0.856 & 0.837 & 0.993 & 1.000 & 1.000 \\
Llama-3.2-1B & 40$\to$80  & 0.855 & 0.856 & 0.855 & 0.899 & 0.855 & 0.999 & 1.000 & 1.000 \\
Llama-3.2-1B & 40$\to$120 & 0.765 & 0.777 & 0.765 & 0.833 & 0.765 & 0.980 & 1.000 & 1.000 \\
Llama-3.2-1B & 80$\to$120 & 0.889 & 0.890 & 0.889 & 0.920 & 0.889 & 1.000 & 1.000 & 1.000 \\
Qwen3-4B FLAN & 80$\to$120 & 0.855 & 0.858 & 0.858 & 0.893 & 0.893 & 0.999 & 1.000 & 1.000 \\
Qwen3-4B 3 ep. & 80$\to$160 & 0.736 & --- & --- & --- & --- & --- & 0.992 & --- \\
Qwen3-4B 3 ep. & 80$\to$240 & 0.702 & --- & --- & --- & --- & --- & 0.985 & --- \\
Qwen3-4B 3 ep. & 80$\to$320 & 0.688 & --- & --- & --- & --- & --- & 0.981 & --- \\
\bottomrule
\end{tabular}
\caption{Top-1,000 overlap with full recomputation for targeted refresh. At $p=0.05$ and $p=0.1$ the columns give the value with (cal.) and without (raw) affine calibration; at $p\ge0.2$ the two agree to within 0.001 on every interval and one value is shown.}
\label{tab:budget_full}
\end{table}

\section{The online check on every single-refresh interval}
\label{app:check}

\begin{table}[H]
\centering
\footnotesize
\setlength{\tabcolsep}{3.5pt}
\begin{tabular}{llccc}
\toprule
Setting & Interval & Check & Stale & Targeted \\
\midrule
Qwen3-4B s0 & 40$\to$80 & 0.641 & 0.631 & 0.931 \\
Qwen3-4B s1 & 40$\to$80 & 0.647 & 0.629 & 0.925 \\
Qwen3-4B s0 & 40$\to$120 & 0.589 & 0.603 & 0.904 \\
Qwen3-4B s1 & 40$\to$120 & 0.533 & 0.568 & 0.913 \\
Qwen3-4B s0 & 80$\to$120 & 0.912 & 0.856 & 1.000 \\
Qwen3-4B s1 & 80$\to$120 & 0.810 & 0.782 & 1.000 \\
Qwen3-4B s2 & 80$\to$120 & 0.876 & 0.817 & 1.000 \\
Qwen3-4B FLAN & 80$\to$120 & 0.911 & 0.855 & 1.000 \\
Llama-3.2-1B & 40$\to$80 & 0.922 & 0.855 & 1.000 \\
Llama-3.2-1B & 40$\to$120 & 0.817 & 0.765 & 1.000 \\
Llama-3.2-1B & 80$\to$120 & 0.958 & 0.889 & 1.000 \\
Qwen3-4B 3 ep. & 80$\to$160 & 0.739 & 0.736 & 0.992 \\
Qwen3-4B 3 ep. & 80$\to$240 & 0.706 & 0.702 & 0.985 \\
Qwen3-4B 3 ep. & 80$\to$320 & 0.687 & 0.688 & 0.981 \\
\bottomrule
\end{tabular}
\caption{The online check on all fourteen single-refresh intervals. Check is the Spearman correlation between stale and recomputed scores on the $p=0.3$ refresh set, observable at refresh time; Stale and Targeted are the top-1,000 overlaps of stale reuse and of targeted refresh at $p=0.3$ with full recomputation.}
\label{tab:check}
\end{table}

\begin{table}[H]
\centering
\scriptsize
\setlength{\tabcolsep}{3.5pt}
\begin{tabular}{llcccccc}
\toprule
Setting & Interval & Global $\rho$ & Stale top-10\% & Check $p{=}0.1$ & Check $p{=}0.2$ & Check $p{=}0.3$ & Band $R^2$ \\
\midrule
Qwen3-4B s0 & 40$\to$80  & 0.911 & 0.631 & 0.526 & 0.621 & 0.641 & 0.433 \\
Qwen3-4B s0 & 80$\to$120 & 0.987 & 0.856 & 0.811 & 0.864 & 0.912 & 0.860 \\
Qwen3-4B s1 & 40$\to$80  & 0.916 & 0.629 & 0.509 & 0.612 & 0.647 & 0.439 \\
Qwen3-4B s1 & 80$\to$120 & 0.976 & 0.782 & 0.745 & 0.781 & 0.810 & 0.717 \\
\bottomrule
\end{tabular}
\caption{The global rank correlation stays above 0.91 on the early interval although the top-10\% overlap drops to 0.63, whereas the check on the refresh set drops to 0.64.}
\label{tab:check_global}
\end{table}

\section{Refresh allocation and iterated chains}
\label{app:chain}

\paragraph{Allocation.}
Table~\ref{tab:allocation_full} lists, for every setting of Table~\ref{tab:fidelity}, the top-1,000 and stratified overlaps of global and per-stratum allocation at $p\in\{0.1,0.2,0.3\}$ together with the source of the examples that the policy misses. Per-stratum allocation spends part of the budget on strata whose scores are low under the target, so its top-1,000 overlap sits below that of global allocation; its target is the stratified subset.

\begin{table}[H]
\centering
\scriptsize
\setlength{\tabcolsep}{3pt}
\begin{tabular}{llcccccc}
\toprule
& & \multicolumn{3}{c}{Global allocation, top-1,000 / stratified} & \multicolumn{3}{c}{Per-stratum allocation, top-1,000 / stratified} \\
\cmidrule(lr){3-5}\cmidrule(lr){6-8}
Setting & stale & $p{=}0.1$ & $p{=}0.2$ & $p{=}0.3$ & $p{=}0.1$ & $p{=}0.2$ & $p{=}0.3$ \\
\midrule
Qwen3-4B s0, \gsm & 0.856 / 0.824 & 0.894 / 0.879 & 0.999 / 0.920 & 1.000 / 0.927 & 0.871 / 0.841 & 0.910 / 0.986 & 0.935 / 0.999 \\
Qwen3-4B s1, \gsm & 0.782 / 0.720 & 0.829 / 0.792 & 0.971 / 0.837 & 1.000 / 0.857 & 0.817 / 0.772 & 0.884 / 0.955 & 0.920 / 0.996 \\
Qwen3-4B s2, \gsm & 0.817 / 0.830 & 0.856 / 0.889 & 0.993 / 0.935 & 1.000 / 0.940 & 0.847 / 0.852 & 0.894 / 0.986 & 0.919 / 0.998 \\
Llama-3.2-1B, \gsm & 0.889 / 0.829 & 0.920 / 0.871 & 1.000 / 0.885 & 1.000 / 0.904 & 0.906 / 0.866 & 0.915 / 0.978 & 0.937 / 0.992 \\
Qwen3-4B, FLAN pool, \gsm & 0.855 / 0.833 & 0.893 / 0.888 & 0.999 / 0.929 & 1.000 / 0.936 & 0.871 / 0.874 & 0.908 / 0.989 & 0.935 / 1.000 \\
Qwen3-4B s0, MMLU & 0.872 / 0.733 & 0.907 / 0.816 & 0.996 / 0.819 & 1.000 / 0.819 & 0.878 / 0.777 & 0.918 / 0.934 & 0.984 / 0.977 \\
Qwen3-4B s1, MMLU & 0.896 / 0.698 & 0.921 / 0.756 & 0.996 / 0.765 & 1.000 / 0.765 & 0.896 / 0.739 & 0.936 / 0.864 & 0.975 / 0.919 \\
\bottomrule
\end{tabular}
\caption{Global against per-stratum refresh allocation at equal budget, checkpoint 80 to 120. Under the \gsm target every example that global allocation misses at $p=0.3$ is a Dolly example (73, 143, 60, 96, and 64 of 1,000 in the five rows); under the MMLU target every missed example is a \gsm example (181 and 235).}
\label{tab:allocation_full}
\end{table}

\paragraph{Iterated chains.}
Table~\ref{tab:chain} gives the post-warmup chain of Figure~\ref{fig:allocation_chain}(b) in full, and Table~\ref{tab:chain_full} lists the chains that start at checkpoint 40, the early part of warmup, for both allocations and both calibrations, and the post-warmup chain with the cohort calibration.

\begin{table}[H]
\centering
\footnotesize
\setlength{\tabcolsep}{3pt}
\begin{tabular}{lccccc}
\toprule
Policy ($p=0.3$) & Cost & Step 160 & Step 240 & Step 320 & Aggregate \\
\midrule
Full recomputation & 4.0 & 1.000 / 1.000 & 1.000 / 1.000 & 1.000 / 1.000 & 1.000 / 1.000 \\
Stale features & 1.0 & 0.736 / 0.573 & 0.702 / 0.625 & 0.688 / 0.586 & 0.820 / 0.713 \\
Iterated, global allocation & 1.9 & 0.992 / 0.749 & 0.984 / 0.816 & 0.979 / 0.786 & 0.999 / 0.844 \\
Iterated, per-stratum allocation & 1.9 & 0.846 / 0.942 & 0.790 / 0.944 & 0.777 / 0.939 & 0.846 / 0.990 \\
\bottomrule
\end{tabular}
\caption{Repeated selection with an iterated cache on the three-epoch run (Qwen3-4B seed 0): a full pass at checkpoint 80, then $0.3N$ features refreshed and written back at each later checkpoint. Cost is the cumulative number of train-gradient features per example. Cells are top-1,000 / stratified overlap with full recomputation at that checkpoint; Aggregate compares the learning-rate-weighted sum of the four checkpoint scores, the LESS selection rule.}
\label{tab:chain}
\end{table} The naive calibration ranks by raw stale score and fits one affine map on the refresh set. The cohort calibration refreshes a uniform random pilot of 5\% of every age cohort of the cache (candidates whose features were last recomputed at the same checkpoint), fits one affine map per cohort on the pilot, ranks the remaining candidates by cohort-calibrated stale score, spends the rest of the budget on that ranking, and refits the cohort maps on the whole refresh set. Check is the Spearman correlation between stale and recomputed scores on the refresh set at that step.

\begin{table}[H]
\centering
\scriptsize
\setlength{\tabcolsep}{2.2pt}
\begin{tabular}{llcccccc}
\toprule
Setting & Policy ($p=0.3$) & Step 1 & Check & Step 2 & Check & Step 3 & Aggregate \\
\midrule
Qwen3-4B s0, 40$\to$80$\to$120 & stale features & 0.631 / 0.456 & --- & 0.603 / 0.477 & --- & --- & 0.782 / 0.664 \\
& global, naive & 0.931 / 0.688 & 0.641 & 0.898 / 0.702 & 0.950 & --- & 0.992 / 0.792 \\
& global, cohort & 0.908 / 0.676 & 0.722 & 0.927 / 0.727 & 0.875 & --- & 0.988 / 0.785 \\
& per-stratum, naive & 0.675 / 0.818 & 0.894 & 0.613 / 0.838 & 0.973 & --- & 0.811 / 0.935 \\
& per-stratum, cohort & 0.676 / 0.791 & 0.904 & 0.751 / 0.875 & 0.944 & --- & 0.844 / 0.912 \\
\midrule
Qwen3-4B s1, 40$\to$80$\to$120 & stale features & 0.629 / 0.473 & --- & 0.568 / 0.394 & --- & --- & 0.775 / 0.649 \\
& global, naive & 0.925 / 0.693 & 0.647 & 0.900 / 0.622 & 0.874 & --- & 0.999 / 0.788 \\
& global, cohort & 0.896 / 0.692 & 0.732 & 0.939 / 0.670 & 0.805 & --- & 0.992 / 0.790 \\
& per-stratum, naive & 0.681 / 0.816 & 0.892 & 0.625 / 0.766 & 0.920 & --- & 0.828 / 0.939 \\
& per-stratum, cohort & 0.679 / 0.795 & 0.902 & 0.768 / 0.856 & 0.900 & --- & 0.861 / 0.930 \\
\midrule
Llama-3.2-1B, 40$\to$80$\to$120 & stale features & 0.855 / 0.668 & --- & 0.765 / 0.566 & --- & --- & 0.911 / 0.765 \\
& global, naive & 1.000 / 0.782 & 0.922 & 1.000 / 0.723 & 0.964 & --- & 1.000 / 0.848 \\
& global, cohort & 1.000 / 0.781 & 0.932 & 1.000 / 0.735 & 0.963 & --- & 1.000 / 0.853 \\
& per-stratum, naive & 0.906 / 0.919 & 0.927 & 0.695 / 0.551 & 0.865 & --- & 0.894 / 0.901 \\
& per-stratum, cohort & 0.898 / 0.910 & 0.943 & 0.861 / 0.881 & 0.812 & --- & 0.946 / 0.958 \\
\midrule
Qwen3-4B s0, 80$\to$160$\to$240$\to$320 & global, cohort & 0.981 / 0.745 & 0.805 & 0.979 / 0.817 & 0.949 & 0.981 / 0.790 & 0.997 / 0.841 \\
& per-stratum, cohort & 0.840 / 0.923 & 0.877 & 0.840 / 0.918 & 0.921 & 0.869 / 0.964 & 0.906 / 0.970 \\
\bottomrule
\end{tabular}
\caption{Iterated chains under the \gsm target: early-origin chains with both calibrations, and the post-warmup chain with the cohort calibration (its naive rows are in Table~\ref{tab:chain} above). Cells are top-1,000 / stratified overlap with full recomputation at that checkpoint; Aggregate compares the learning-rate-weighted checkpoint sums. Under the MMLU target the early-origin global chains reach 0.891 and 0.929 top-1,000 overlap at checkpoint 120 and 0.960 and 0.973 for the aggregate on seeds 0 and 1.}
\label{tab:chain_full}
\end{table}

\section{Selection rules}
\label{app:rules}

\begin{table}[H]
\centering
\footnotesize
\setlength{\tabcolsep}{4pt}
\begin{minipage}[t]{0.50\textwidth}
\centering
\begin{tabular}{lcc}
\toprule
Rule (GSM200, seed 0) & Full & $p=0.3$ \\
\midrule
Unconstrained top-$k$ & 0.220 & --- \\
Length quotas & 0.100 & 0.160 \\
Source quotas & 0.410 & 0.440 \\
Source $\times$ length quotas (50/50) & 0.655 & 0.600 \\
Source $\times$ length quotas (70/30) & 0.400 & 0.440 \\
Source $\times$ length quotas (80/20) & 0.350 & 0.345 \\
BIDS-style, source & 0.535 & 0.540 \\
BIDS-style, source $\times$ length & 0.615 & 0.625 \\
\bottomrule
\end{tabular}
\end{minipage}\hfill
\begin{minipage}[t]{0.48\textwidth}
\centering
\begin{tabular}{lcc}
\toprule
Rule (full \gsm, seed 0) & Full & $p=0.3$ \\
\midrule
Source $\times$ length quotas (labels) & 0.661 & 0.616 \\
Cluster $\times$ length, proportional & 0.579 & 0.468 \\
Cluster $\times$ length, uniform & 0.350 & 0.406 \\
Plain random & 0.529 & --- \\
Unconstrained top-$k$ & 0.196 & --- \\
\bottomrule
\end{tabular}
\end{minipage}
\caption{Selection rules on seed 0. Left: single-run exact match on the 200-item GSM200 probe for quota, ratio, and normalization variants. Right: single-run full \gsm exact match for the label-free cluster rule ($k$-means with $k=8$ on cached features) against the labeled rule and controls; Table~\ref{tab:noise} gives the training-run range for single runs.}
\label{tab:rules}
\end{table}

\section{Cost accounting}
\label{app:cost}

\begin{table}[H]
\centering
\footnotesize
\setlength{\tabcolsep}{4pt}
\begin{tabular}{lrrl}
\toprule
Stage & Full recompute & \method $p=0.3$ & Changed \\
\midrule
Train-gradient features at checkpoint 120 & 3{,}935.75\,s & 1{,}087.33\,s & yes \\
Validation-gradient extraction & 106.70\,s & 106.70\,s & no \\
Scoring, calibration, selection (CPU) & 6.83\,s & 6.83\,s & no \\
\midrule
Selection cycle & 4{,}049.28\,s & 1{,}200.86\,s & 3.37$\times$ \\
\midrule
LoRA training on 1,000 selected examples & 1{,}433\,s & 1{,}442\,s & no \\
\bottomrule
\end{tabular}
\caption{Wall-clock accounting of one selection cycle on Qwen3-4B-Base with the 10k seed-0 pool on one NVIDIA A6000. The train-gradient stage covers 10,000 examples for full recomputation and 3,000 for \method; LoRA times are two-seed means. The cache footprint is $10{,}000\times 8{,}192$ float32 values, about 313\,MB.}
\label{tab:cost}
\end{table}

All inputs of the amortized scenario are the measured values of Table~\ref{tab:cost}. Over $R$ selection cycles the saving is $R\cdot 4049/(4049+(R-1)\cdot 1201)$, since \method pays one full pass to build the cache. The pool-size extrapolation scales the train-gradient stage with the pool and keeps validation gradients, materialization, and training fixed.

\begin{table}[H]
\centering
\footnotesize
\setlength{\tabcolsep}{4pt}
\begin{minipage}[t]{0.40\textwidth}
\centering
\begin{tabular}{rcc}
\toprule
Cycles $R$ & Selection stage & With training \\
\midrule
1 & 1.00$\times$ & 1.00$\times$ \\
2 & 1.54$\times$ & 1.35$\times$ \\
3 & 1.88$\times$ & 1.53$\times$ \\
5 & 2.29$\times$ & 1.71$\times$ \\
10 & 2.73$\times$ & 1.87$\times$ \\
$\infty$ & 3.37$\times$ & 2.07$\times$ \\
\bottomrule
\end{tabular}
\end{minipage}\hfill
\begin{minipage}[t]{0.56\textwidth}
\centering
\begin{tabular}{rccc}
\toprule
Pool & Selection cycle & With training & $R=5$, cache paid \\
\midrule
10k & 3.37$\times$ & 2.07$\times$ & 2.29$\times$ \\
50k & 3.57$\times$ & 3.04$\times$ & 2.36$\times$ \\
100k & 3.59$\times$ & 3.29$\times$ & 2.37$\times$ \\
\bottomrule
\end{tabular}
\end{minipage}
\caption{Left: speedup over $R$ selection cycles when \method pays one full pass to build the cache. Right: steady-state per-cycle speedup as the pool grows.}
\label{tab:amortized}
\end{table}

\section{Subset transfer to a third model family}
\label{app:mistral}

To test whether the cached subset behaves like the recomputed subset when a different model consumes it, we train Mistral-7B-v0.3 on the seed-0 subsets selected by Qwen3-4B with the stratified rule under full recomputation and under $p=0.3$, plus a plain-random control, without any Mistral-side gradient computation. The two stratified subsets reach 0.397 and 0.396 full \gsm exact match on Mistral, so the difference between cached and recomputed selection stays negligible under transfer. Plain random reaches 0.427, above both influence-selected subsets, because those subsets were chosen with Qwen3-4B gradients and carry the length profile of Qwen3-4B's preferences; for Mistral they are a fixed, source-balanced sample selected by another model, and the diagnostic isolates the cache from the selector. Table~\ref{tab:fidelity} reports the within-model result for the second family.

\begin{table}[H]
\centering
\small
\begin{tabular}{lcc}
\toprule
Transferred subset (selected by Qwen3-4B, seed 0) & Mistral train loss & Mistral full \gsm EM \\
\midrule
Plain random & 0.740 & 0.427 \\
Stratified, full recompute & 0.619 & 0.397 \\
\method, $p=0.3$ & 0.649 & 0.396 \\
\bottomrule
\end{tabular}
\caption{Subset transfer to Mistral-7B-v0.3.}
\label{tab:mistral}
\end{table}

\section*{AI use statement}
In this work, we used generative AI tools (a large language model assistant) for writing support: editing the manuscript text for grammar and clarity, proofreading, checking that terminology and notation stay consistent across sections, and formatting references. We have reviewed all AI-assisted work: the authors read every edited passage against the original wording, checked that each sentence states the result recorded in the experiment outputs, and verified the formatted references against their sources. We take responsibility for the final content of this work, including text, claims, and artifacts produced with the aid of generative AI.

\section*{Reproducibility statement}
Appendix~\ref{app:impl} lists every hyperparameter of warmup, gradient collection, selection, downstream training, and evaluation, and Appendix~\ref{app:prompts} gives the prompt and answer formats. The code release contains the refresh and calibration implementation, the stratified, normalized, and cluster-based selection rules, the global and per-stratum allocation and iterated-cache simulations, the paired bootstrap and training-seed decomposition tools, the refresh-budget and band-agreement analysis scripts, the patch that adapts the public LESS pipeline, and the run scripts for every table in this paper. All datasets are public.

\section*{Ethics statement}
This work studies data selection for post-training language models on public datasets and introduces no human-subject data. Cheaper selection lowers the cost of targeted post-training, and our results show that influence scores applied without diversity constraints select a narrow subset that generalizes poorly, which argues for evaluating selected data on coverage and transfer as well as target accuracy.

\end{document}